\documentclass[sigconf]{acmart}
\AtBeginDocument{%
  \providecommand\BibTeX{{%
    \normalfont B\kern-0.5em{\scshape i\kern-0.25em b}\kern-0.8em\TeX}}}

\copyrightyear{2022}
\acmYear{2022}
\setcopyright{acmcopyright}
\acmConference[KDD '22] {Proceedings of the 28th ACM SIGKDD Conference on Knowledge Discovery and Data Mining}{August 14--18, 2022}{Washington, DC, USA.}
\acmBooktitle{Proceedings of the 28th ACM SIGKDD Conference on Knowledge Discovery and Data Mining (KDD '22), August 14--18, 2022, Washington, DC, USA}
\acmPrice{15.00}
\acmDOI{10.1145/3534678.3539098}
\acmISBN{978-1-4503-9385-0/22/08}

\usepackage[utf8]{inputenc} 
\usepackage[T1]{fontenc}    
\usepackage{hyperref}       
\usepackage{url}            
\usepackage{booktabs}       
\usepackage{amsfonts}       
\usepackage{nicefrac}       
\usepackage{microtype}      
\usepackage{xcolor}         

\usepackage{natbib}
\usepackage{graphicx}
\usepackage{bm}
\usepackage{multirow}
\usepackage{amsmath}
\usepackage{subfigure}
\usepackage{wrapfig}
\usepackage{booktabs}
\usepackage{caption}
\usepackage{algorithm}  
\usepackage{algorithmicx}  
\usepackage{algpseudocode}
\usepackage{threeparttable}

\begin{document}
\title{Multi-task Envisioning Transformer-based Autoencoder for Corporate Credit Rating Migration Early Prediction}

\author{Han Yue}
\email{hanyue@brandeis.edu}
\affiliation{%
  \institution{Brandeis University}
  \city{Waltham}
  \state{Massachusetts}
  \country{USA}
}

\author{Steve Xia}
\email{steve_xia@glic.com}
\affiliation{%
  \institution{Guardian Life Insurance}
  \city{New York}
  \state{New York}
  \country{USA}
}

\author{Hongfu Liu}
\email{hongfuliu@brandeis.edu}
\affiliation{%
  \institution{Brandeis University}
  \city{Waltham}
  \state{Massachusetts}
  \country{USA}
}


\begin{abstract}
Corporate credit ratings issued by third-party rating agencies are quantified assessments of a company's creditworthiness. Credit Ratings highly correlate to the likelihood of a company defaulting on its debt obligations. These ratings play critical roles in investment decision-making as one of the key risk factors. They are also central to the regulatory framework such as BASEL II in calculating necessary capital for financial institutions. Being able to predict rating changes will greatly benefit both investors and regulators alike. In this paper, we consider the corporate credit rating migration early prediction problem, which predicts the credit rating of an issuer will be upgraded, unchanged, or downgraded after 12 months based on its latest financial reporting information at the time. We investigate the effectiveness of different standard machine learning algorithms and conclude these models deliver inferior performance. As part of our contribution, we propose a new Multi-task Envisioning Transformer-based Autoencoder (META) model to tackle this challenging problem. META consists of Positional Encoding, Transformer-based Autoencoder, and Multi-task Prediction to learn effective representations for both migration prediction and rating prediction. This enables META to better explore the historical data in the training stage for one-year later prediction. Experimental results show that META outperforms all baseline models. 
\end{abstract}



\maketitle

\section{Introduction}
    Issuer credit ratings, developed by Moody's in 1914 and by Poor's Corporation in 1922, are one of the most important indicators of a corporation's credit quality. After the issuance and assignment of the initial bond rating, these agencies regularly perform reviews of the underlying issues, which may result in a change in the rating being upgraded or downgraded.  \citet{altman1998importance,saunders2010credit,hardle2017applied} point out that credit rating migration plays an integral part in the more general field of credit risk assessment of corporate bonds. Thus, corporation ratings or rating migration is one of the most crucial factors in investment decisions. Credit ratings are also a critical part of the modern financial regulatory framework because a regulated financial institution's required capital (RBC, or risk-based capital) is directly linked to the qualify of its assets, as measured by credit ratings.
    
    Rating changes often have material impacts on the performance of an issuer's debt. Being able to predict rating changes ahead of time has many applications, from better returns for an investor to the ability to better control and monitor risks for regulators. In this paper, we consider addressing the corporate credit rating migration early prediction challenge, which predicts whether the credit rating of a corporate issuer will be upgraded, remain unchanged, or downgraded after a period of time, such as 12 months. 
    
    In recent years, machine learning techniques are becoming popular in the financial area. For example, \citet{ahelegbey2019latent} and \citet{chen2020encoding} use advanced machine learning and deep learning techniques to study financial risks. \citet{guo2008neural}, \citet{puh2019detecting}, and \citet{nur2020comparative} explore the usage of machine learning for credit card fraud detection. \citet{liebana2017sem} ,\citet{liebana2018predicting}, and \citet{hew2019age} build machine learning models in the banking operations domain. \citet{villuendas2017naive} and \citet{xia2018novel} work towards credit scoring. \citet{lee2007application} and \citet{jabeur2020machine} adopt machine learning methods for bond rating prediction. In this paper, we focus on the rating migration early prediction problem, that is, to forecast the migration of the bond rating of a corporate before it happens. To the best of our knowledge, there are no designated machine learning techniques for rating migration early prediction.
    
    
    The rating migration early prediction problem is quite challenging. One of the main reasons is corporate credit ratings, issued by third-party rating agencies, are evaluated by taking many macro and issuer-specific factors into consideration. The process is impacted by both quantitative and qualitative considerations and sometimes complicated by the business relationships between the rating agencies and the issuers. Moreover, different rating agencies may have different opinions on the same issuer at any specific point in time. In this paper, we focus on solving the following two challenges: the first one is how to find the pattern hidden behind the highly dynamic corporate fundamental time-series data;  the second challenge lies in early prediction. Since we aim to predict the rating migration of a period of time later, the corporate data during such a period are unavailable to be used for prediction, which further increases the difficulty.

    
    
    To tackle the two challenges mentioned above, we propose Multi-task Envisioning Transformer-based Autoencoder (META) to address the rating migration early prediction problem. META consists of Positional Encoding, Transformer-based Autoencoder, and Multi-task Prediction to learn effective representations for both migration prediction and rating prediction. Specifically, we first adopt a Positional Encoding layer, which incorporates time information with corporate information to handle the time-series issue. Then we design a Transformer-based Autoencoder to learn envisioning ability from historical data, thus addressing the second challenge. Finally, we apply Multi-task Prediction to improve the migration task by performing a rating prediction task simultaneously. In summary, we highlight our contributions as follows:\vspace{-1mm}
    \begin{itemize}
        \item We consider a novel problem, corporate credit rating migration early prediction, which predicts the credit rating of a corporate will be upgraded, unchanged, or downgraded after a period of time. To our best knowledge, this crucial problem has not been explored in the literature. 
        \item We propose a Multi-task Envisioning Transformer-based Autoencoder (META) model, which consists of Positional Encoding, Transformer-based Autoencoder, and Multi-task Prediction to learn effective representations for both migration prediction and rating prediction.
        \item Technically, we address the early prediction problem of the time gap between the training and prediction timestamp, where the corporate data during the gap period are unavailable in the training stage.
        \item Extensive experimental results demonstrate that our proposed META performs better than other baseline models. We also provide in-depth explorations of META with more interpretation in prediction tasks and the time gap between training data and test data.
    \end{itemize}
\vspace{-1mm}    
\section{Related Work}
In this section, we introduce the related work in terms of machine learning applications in the finance domain and techniques in time-series prediction and highlight the difference between our research problem and the early prediction in the literature.

\noindent\textbf{Machine Learning in Financial Area}. In the past decade, applications of machine learning techniques in finance have become been steadily increasing \cite{sezer2020financial,cao2021data,hendershott2021fintech}. Among diverse applications in finance, the prediction problem is one of the most common. Stock price perdition is one of the most studied financial applications. \citet{vargas2017deep} adopt a model composed of Convolutional Neural Network (CNN) and Long-Short Term Memory network (LSTM) to extract information from S\&P500 index news for price prediction and intraday directional movement estimation. \citet{das2018real} use a Recurrent Neural Network (RNN) together with sentiment analysis on Twitter for stock price forecasting. \citet{zhou2018stock} use a Generative Adversarial Network to minimize forecast error loss and direction prediction loss for stock price prediction. Some studies focus on predicting stock market indexes instead of prices. \citet{dingli2017financial} make use of CNN to forecast the next period direction of S\&P500 index. \citet{jeong2019improving} build a Reinforcement Learning model to predict S\&P500 index. LSTM-based models with various other data \cite{si2017multi,mourelatos2018financial,chen2018stock} are developed for index prediction. Some researchers also focus on forecasting the underlying volatility of assets, where volatility is a statistical measure of the dispersion of returns for risk assessment and asset pricing. \citet{doering2017convolutional} implement a CNN model for volatility prediction. \citet{zhou2019long} use LSTM and keywords from daily search volume based on Baidu to predict index volatility. Others~\cite{nikolaev2013time,psaradellis2016modelling,kim2018forecasting} design generalized autoregressive conditional heteroscedasticity (GARCH)-type models for volatility prediction. There are also several studies on bond rating prediction. \citet{lee2007application} apply Support Vector Machine (SVM), and \citet{kim2012ensemble} use ensemble learning to predict bond rating. \citet{jabeur2020machine} adopt the cost-sensitive
decision tree algorithm~\cite{chauchat2001targeting} as well as SVM and Multi-Layer Perceptron (MLP) for bond rating prediction. While machine learning methods are widely applied in finance, the early prediction of rating migrations has not been explored as far as we know. 

\noindent\textbf{Time-Series Prediction}. Time-series predicting plays an essential role in a wide range of real-life problems, thus leading to a huge body of works in this area. Many Recurrent Neural Network (RNN)-based methods~\cite{rangapuram2018deep,wang2019deep,lim2020recurrent,salinas2020deepar} have been developed for time-series prediction by modeling the temporal dependencies. LSTM~\cite{hochreiter1997long} is one of the most popular RNNs, which addresses the problem of exploding and vanishing gradients by improving gradient flow with a cell, an input gate, an output gate, and a forget gate. Gated recurrent units~\cite{cho2014learning} is similar to LSTM but without an output gate. DeepAR~\cite{salinas2020deepar} combines autoregressive with RNNs and produces a probabilistic distribution of time series. Inspired by CNNs, which can capture local relationships, some researchers also design causal convolutions that use past information for forecasting. WaveNet~\cite{oord2016wavenet} uses CNN to represent the conditional distribution of the acoustic features given the linguistic feature. TCN~\cite{bai2018empirical} combines residual connections with causal convolutions. LSTNet~\cite{lai2018modeling} uses CNN with recurrent-skip connections to capture temporal patterns. Recently, Transformer~\cite{vaswani2017attention}-based architectures using attention mechanisms show great power in sequential data. \citet{fan2019multi} use attention to aggregate features extracted by Bi-LSTM. LogTrans~\cite{li2019enhancing} adopts causal convolution to capture local context and designed LogSparse attention to select time steps. Informer~\cite{zhou2021informer} extends Transformer with KL-divergence-based self-attention mechanism, distilling operation, and a generative style decoder.

Different from the methods mentioned above, we focus on a novel time-series prediction approach, which is predicting several steps ahead instead of predicting the following step immediately. Although early prediction problems in the data mining area have been addressed in several scenarios, including disease diagnosis~\cite{zhao2019asynchronous,xing2009early}, student performance prediction~\cite{chen2020utilizing,raga2019early}, action recognition~\cite{kong2017deep,tran2021progressive}, and so on, these problems usually consider the time-series data that only access partial or preceding frames to recognize its static category along with all the timestamps. On the contrary,  the problem we address here has dynamic labels, i.e., the same corporate credit ratings at different timestamps might be different.

\begin{figure*}[t]
        \centering
        \subfigure[Number of Companies with different ratings]{
                \begin{minipage}[t]{0.31\textwidth}
                \centering
                \label{fig:stat_rating_num}
                \includegraphics[scale=0.15]{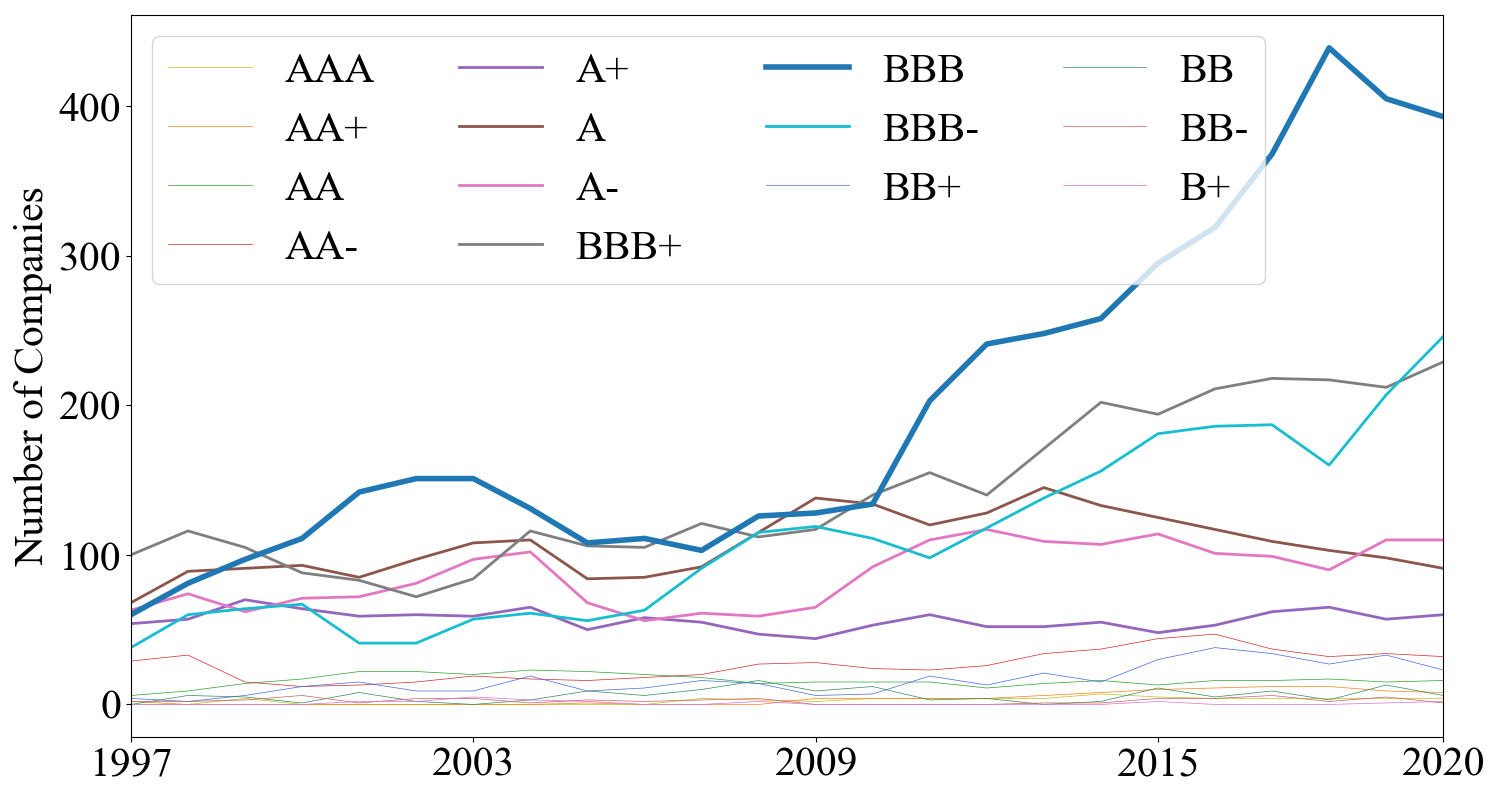}
                \end{minipage}
            }
        \subfigure[Percentage of Companies with different ratings]{
                \begin{minipage}[t]{0.31\textwidth}
                \centering
                \label{fig:stat_rating_per}
                \includegraphics[scale=0.15]{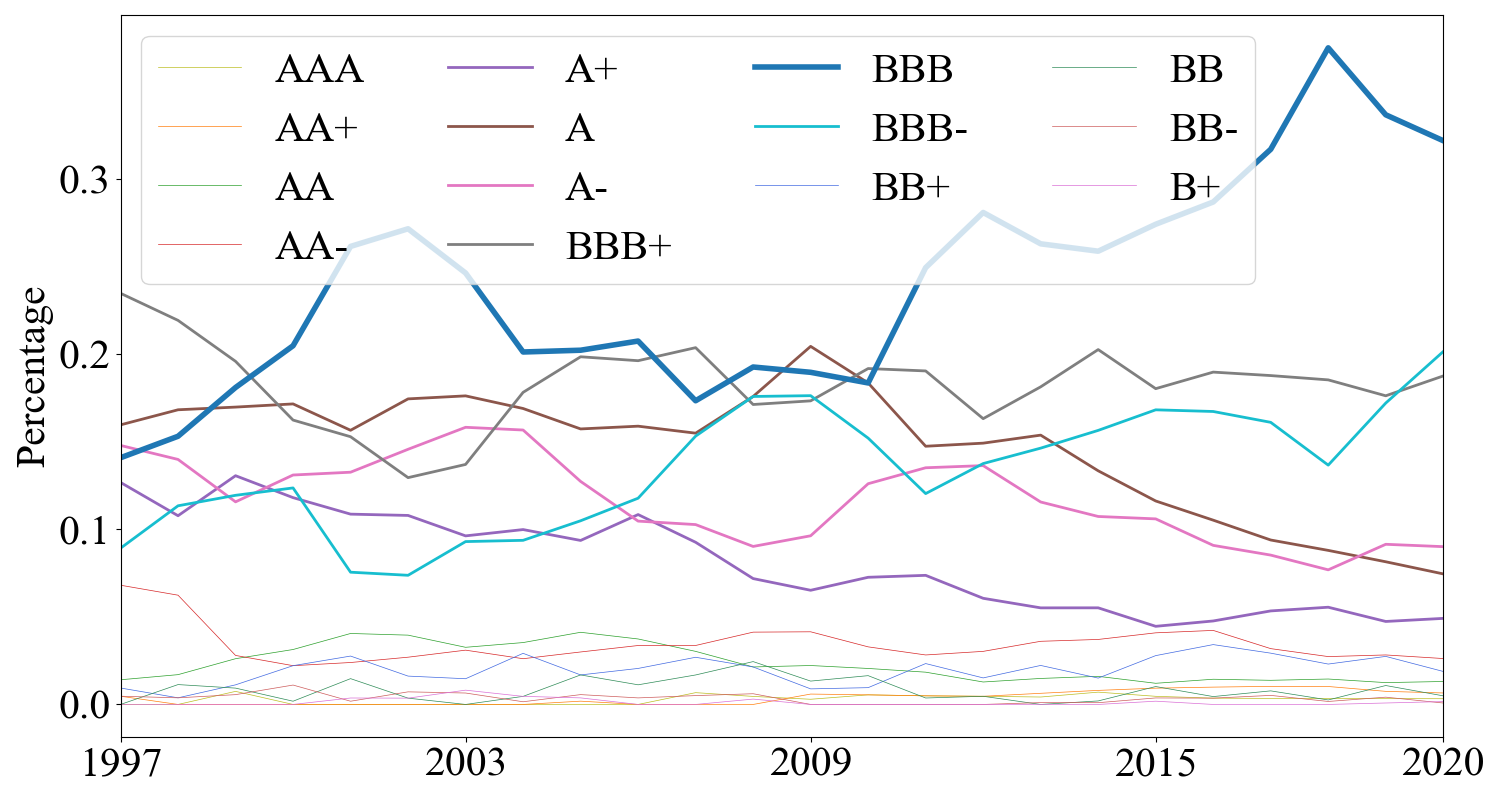}
                \end{minipage}
            }
        \subfigure[Upgrade/Downgrade percentage]{
                \begin{minipage}[t]{0.31\textwidth}
                \centering
                \label{fig:stat_up_down}
                \includegraphics[scale=0.15]{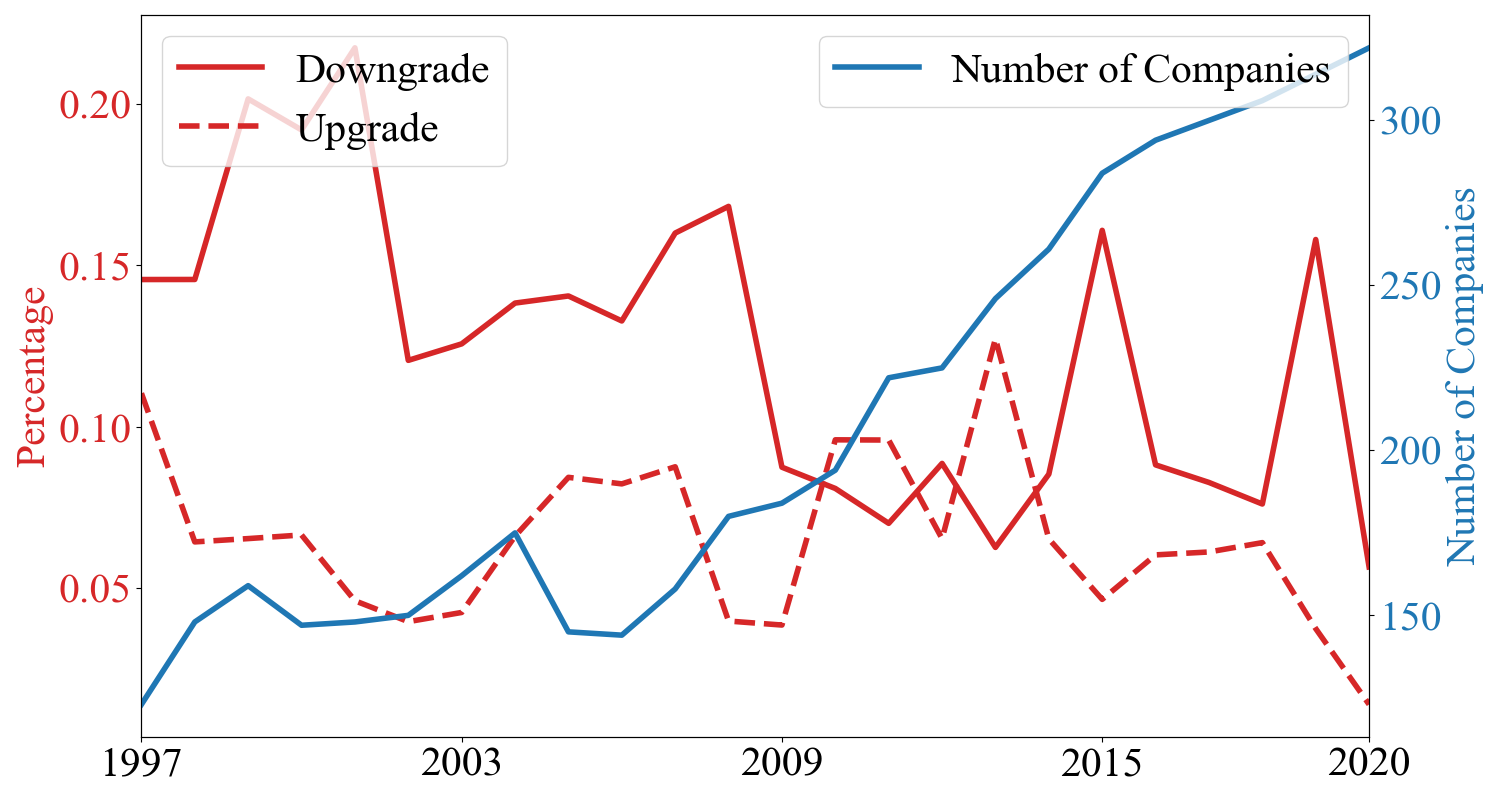}
                \end{minipage}
            }
        
        \subfigure[Balance sheet features]{
                \begin{minipage}[t]{0.31\textwidth}
                \centering
                \label{fig:stat_bs}
                \includegraphics[scale=0.15]{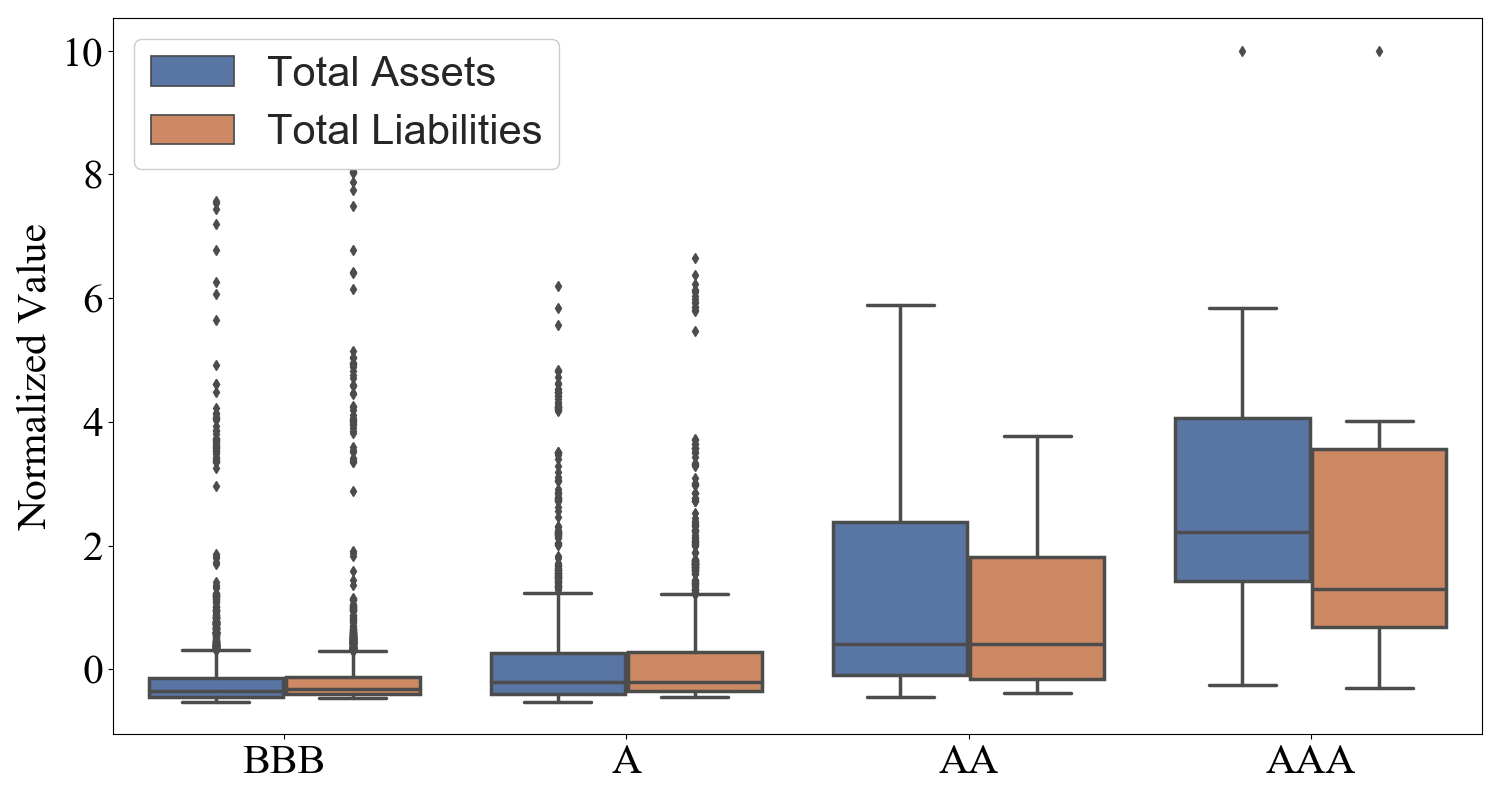}
                \end{minipage}
            }
        \subfigure[Market features]{
                \begin{minipage}[t]{0.31\textwidth}
                \centering
                \label{fig:stat_md}
                \includegraphics[scale=0.15]{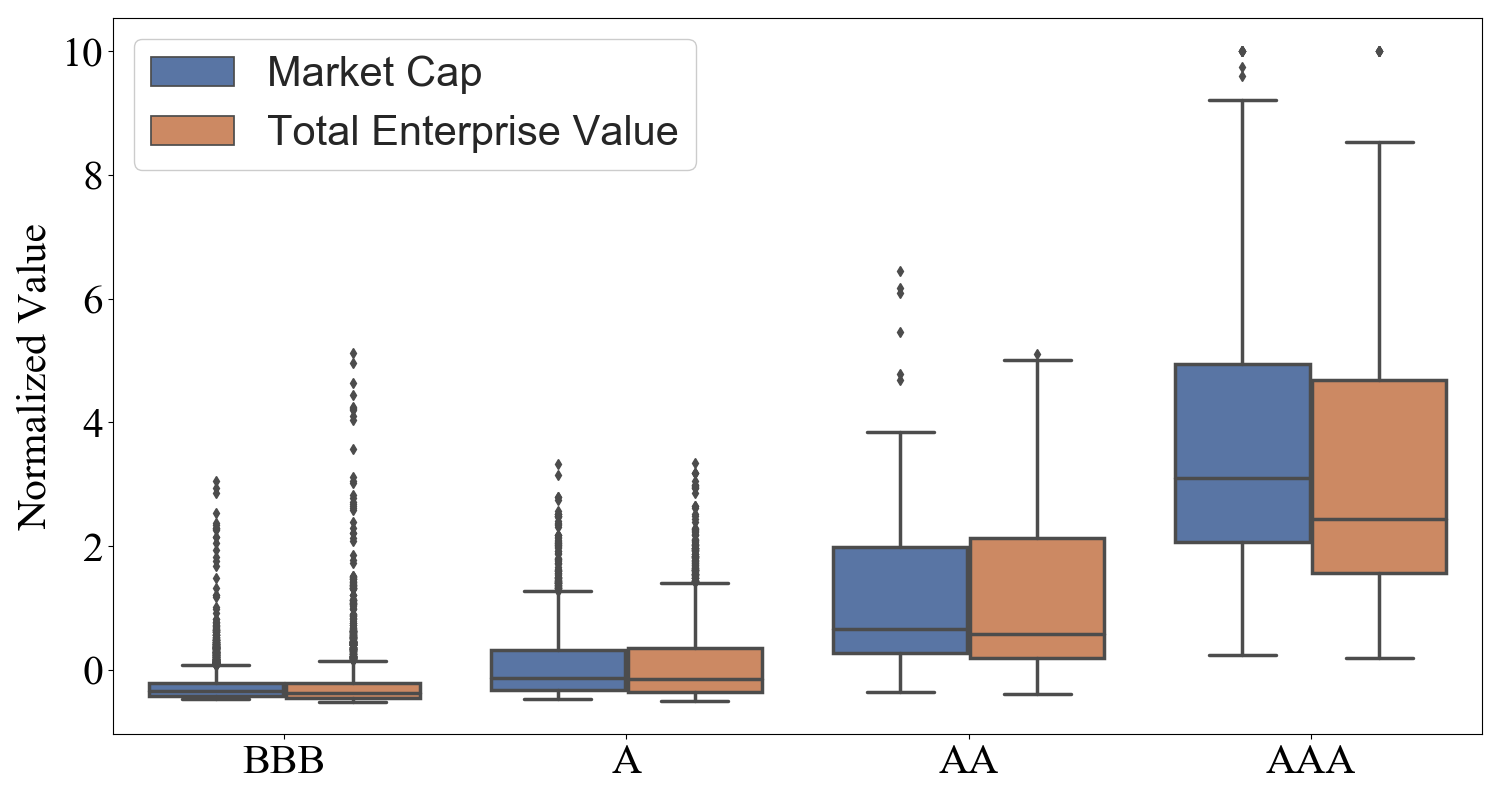}
                \end{minipage}
            }
        \subfigure[Rating migration matrix]{
                \begin{minipage}[t]{0.31\textwidth}
                \centering
                \label{fig:stat_matrix}
                \includegraphics[scale=0.15]{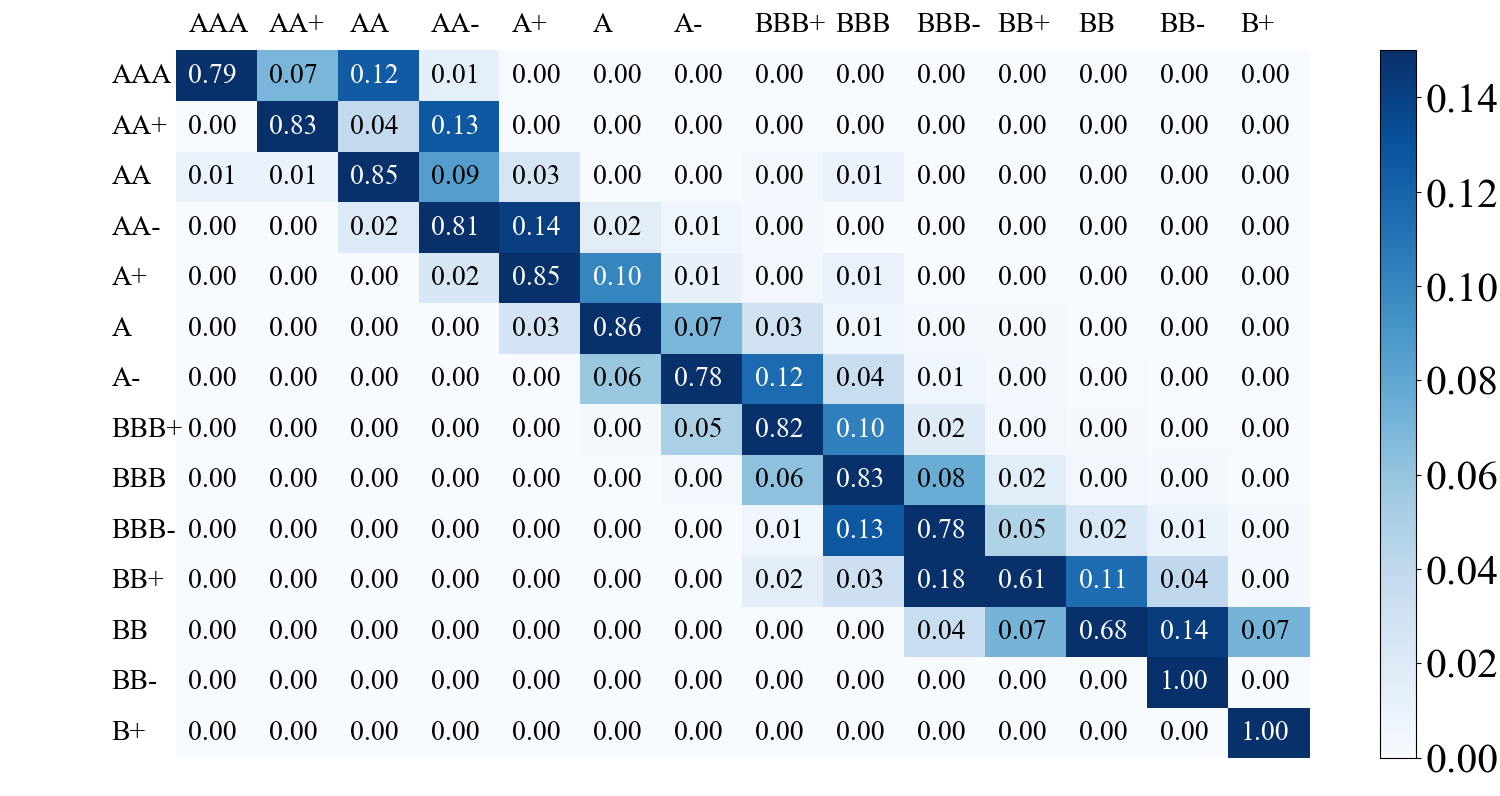}
                \end{minipage}
            }
        \vspace{-2mm}
        \caption{Statistics of ratings and companies. (a-b) Number and percentage of companies with 14 ratings from AAA to B+ by year from 1997 to 2020; (c) Percentage of upgrade and downgrade companies and the total number of companies in the same period; (d-e) Distributions of four corporate features in balance sheet and market among 4 rating groups in AAA, AA, A and BBB; (f) Rating migration heatmap among different rating groups.}
        \label{fig:stat}
        \vspace{-2mm}
    \end{figure*}

\section{Data Description}

    In this paper, we use a dataset of historical commercial corporate ratings with a time length of 24 years ranging between 1997 and 2020. It includes a total of 445 unique companies. Data used contains two parts: corporate information and rating migrations, which we provide more details below. These data came from rating agencies.
    
    \textbf{Corporate information}. Indexed by filing date and corporate identifier, the corporate information data contain 70 features, including 8 features in balance sheet, 2 features in capital structure, 3 features in cash flow, 8 features in income statement, 5 features in market data, 3 supplemental features, 6 features based on feature engineering, and 32 features based on ratios from long-term liquidity, short-term liquidity, profitability, margin analysis, and above-mentioned feature types. The interval of data for a same company is about 3 months in most cases. 
    
    \textbf{Rating migrations}. The ratings of companies in the dataset include 18 levels, ranging from high to low are: `AAA,' `AA+,' `AA,' `AA-,' `A+,' `A,' `A-,' `BBB+,' `BBB,' `BBB-,' `BB+,' `BB,' `BB-,' `B+,' `B,' `B-,' `D,' `NR.' There are two kinds of rating migration involved in the dataset: downgrade and upgrade. The downgrade means the rating of a company is changed to a lower level (higher risk), and the upgrade means the opposite. Different from corporate information data, the rating migration data is not seasonal due to the fact that rating agencies can change the rating at any time. Overall, there are 3384 rating migrations that happened during the time period, where 1964 of them are downgrades, and 1420 are upgrades.
    
    \textbf{Data pre-processing}. To keep the dates and identifiers of both parts of the data consistent for usage, we match the ratings of corporates with the indices of the corporate information data and calculate a 12-month rating migration for each corporate based on the rating migration data in the next 12 months. We also go through several further steps for data pre-processing. Firstly, we normalize the data to eliminate the influence of feature magnitude. Secondly, we fill in zeros for the missing data to guarantee the functioning of algorithms. Thirdly, we remove data points with ratings of `B,' `B-,' and `D' due to that the number of their migrations is less than 10. We also remove data points with ratings of `NR' because the corresponding companies are not rated by rating agencies. Finally, we re-organize the corporate information into time-series data with an interval of 3 months (seasonal) based on corporate identifiers. For those who do not have enough previous data, we use the closest data point in time instead to make sure that all data points have the same dimension. After pre-processing, we get a total of 18674 data points with 14 levels of ratings, where the numbers of upgraded, downgrades, and unchanged are 1175, 2164, and 15335, respectively.
    
    \textbf{Visualization of statistics}. Figure~\ref{fig:stat} shows statistics of the processed data. Figure~\ref{fig:stat_rating_num} and Figure~\ref{fig:stat_rating_per} plot the number and percentage of companies with different ratings every year, respectively. `BBB' is the most common rating among companies almost every year. A majority of companies are rated between `BBB-' and `A+.' Figure~\ref{fig:stat_up_down} shows how the number of companies changes and the total number of companies during the period. The data of a company in the dataset may only cover part of the 24-year window and is not available for the rest of the time due to activities such as Merger \& Acquisition, bankruptcy and going private, etc. Thus the number of companies varies every year. In general, the total number of companies increases over the years, indicating new companies join the market. Especially from 2006, the number of companies keeps increasing significantly every year. Meanwhile, the percentages of upgrades and downgrades do not change much and are less than 5\%. The number of companies that received a downgrade is always more than that of companies upgraded except for 2010 and 2013, which might be because of the economic recovery after a recession. We also demonstrate the distributions of total assets, total liabilities, market cap, and total enterprise value in balance sheet and market among 4 rating groups in AAA, AA, A, and BBB in Figure~\ref{fig:stat_bs}~\&~\ref{fig:stat_md}. The companies in AAA have much higher values in terms of the above four features than ones in other categories. The companies rated BBB have lower average but higher variances in these four features relative to higher-rated companies. Figure~\ref{fig:stat_matrix} shows the rating migration heatmap among 14 rating groups. The numbers in the dialog mean the unchanged ratios, which indicates the rating migration problem is an extremely imbalanced prediction problem. While considering the sizes of different rating groups, most rating migrations occur in A+, A, A-, BBB+, BBB, and BBB-. 
 
\section{Problem Formulation}
    \label{sec:problem}
    
    Rating migration is more relevant to long-term trading for investors compared to short-term trading. Given historical information on companies, the goal of the rating migration prediction problem is to predict how the rating of these companies will change after 12 months. The migration includes upgraded, downgraded, and unchanged. Thus it can be formulated as a multiclass classification problem with 3 categories. Let $T$ denote the time range of historical data, $D$ the dimension of corporate information, $X \in \mathbb{R}^{T \times D}$ the time-series information of a corporate, and $Y \in \{-1,0,1\}^{T}$ the corresponding rating migrations, and then the problem can be formulated as to find a mapping function $f:X^N \to Y^N$, where $N$ is the number of samples, such that a rating migration is predicted given historical data of a corporate.
    
    In this prediction problem, there are two challenges we want to deal with. The first one is about the time-series nature of rating change decisions. The rating of a corporate is changed based on information available to the rating agencies with a strong historical perspective. This means that historical company performance, as well as the variances of its performance between different dates, have an impact on whether and how the rating will be changed. Therefore, it is necessary to capture the hidden momentum from historical time-series data in developing a model. The second challenge is lagged training data. Because our goal is to predict the rating migration of companies after 12 months, the corporate information data used for model training must be at least 1 year ago or even earlier to ensure the availability of labeling. It means that we are trying to predict the rating migrations of companies with the most recent data based on somewhat outdated data. 
    
\section{Methodology}
    \label{sec:method}
    In this section, we first introduce the framework of our proposed model, then elaborate on the objective function in detail.

    \subsection{Framework of META}
    
        \begin{figure*}[t]
            \centering
            \includegraphics[scale=0.48]{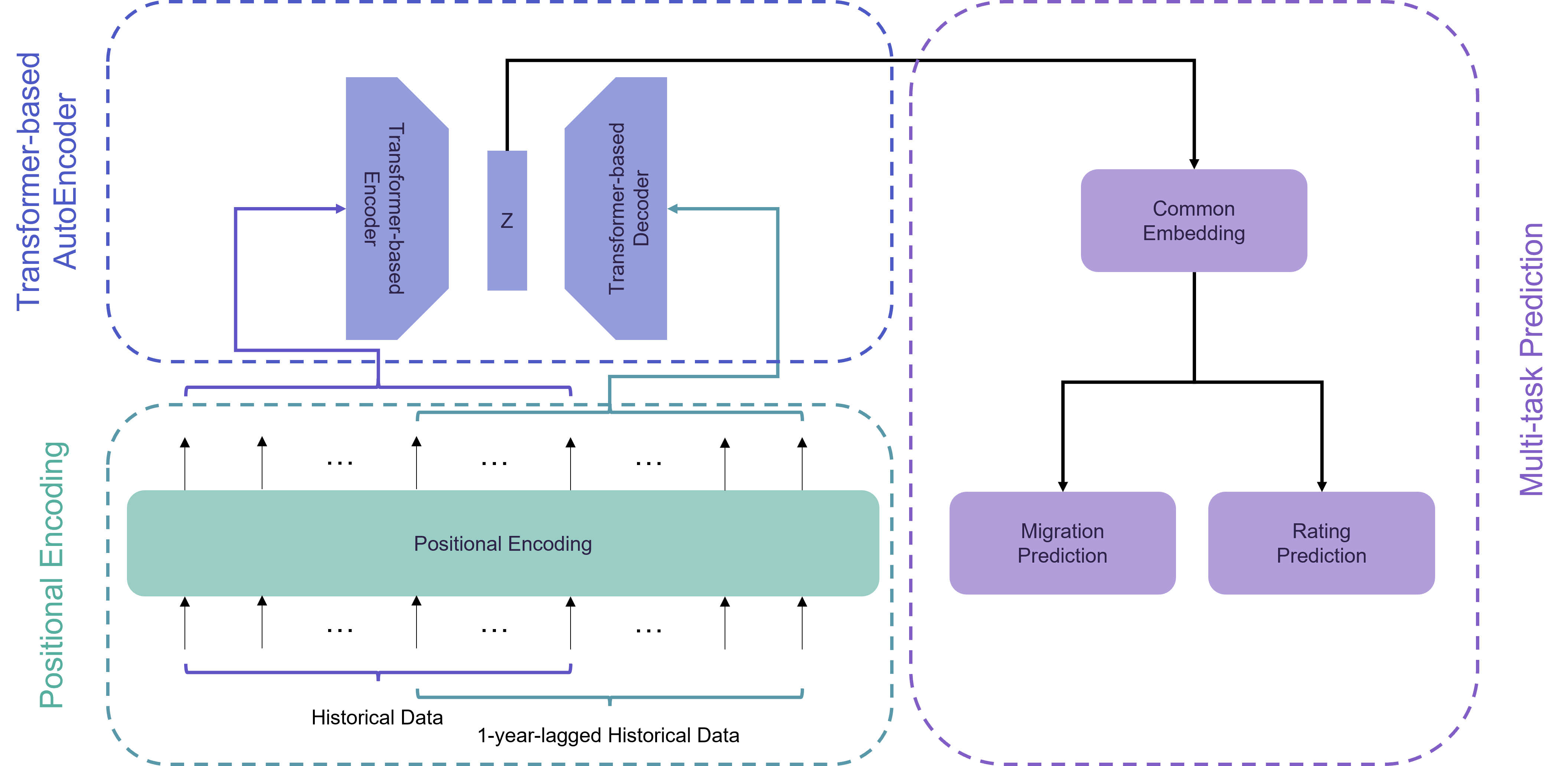}\vspace{-2mm}
            \caption{Framework of Multi-task Envisioning Transformer-based Autoencoder (META). Our proposed META consists of three components: Positional Encoding, Transformer-based Autoencoder, and Multi-task Prediction. The model takes two kinds of features as input. The first one is historical data available at a specific time, and the other one is the corresponding 1-year-later data. The second kind of data is only used in the model training stage. The Positional Encoding part takes these time-series data as input and generates embeddings with position information to handle the sequential manner. The Transformer-based Autoencoder is designed to translate the current data into 1-year-lagged data, where the hidden features with this envisioning ability are taken out for the next step. Finally, in the Multi-task Prediction part, a Common Embedding layer is built to get representations from the hidden features of the Transformer-based Autoencoder. Based on the common embeddings, a Migration Prediction layer, as well as a Rating Prediction layer, are built to get predictions for migrations and ratings.}
            \label{fig:model}
            \vspace{-4mm}
        \end{figure*}
        
        We propose our model to address the challenges mentioned in Section~\ref{sec:problem}. Figure~\ref{fig:model} shows the framework of META, which mainly consists of three parts: Positional Encoding, Transformer-based Autoencoder, and Multi-task Prediction. META takes inputs from two periods, where data from the first period is what we want to predict migrations for, and data from the second period is the 1-year-lagged data of the first period. In order to handle the time-series historical data, we adopt Positional Encoding to generate representations with time information included for inputs of both periods. Then we build a Transformer-based Autoencoder to take representations of the first period as input, and we use Mean Squared Error to align the decoding output with representations of the second period, giving META the ability to envision 1-year-lagged data. While the 1-year-lagged data is only required in the model training stage and not needed in the test stage, all the historical data is used either as the first-period or as the second-period data in model training, and no extra data is required for the model test. Thus the issue of outdated data is addressed. Considering that the migration prediction task has a close relationship with the rating prediction task, the learning of the two tasks may help each other. Therefore, in the Multi-task Prediction part, we build a Common Embedding layer on top of the hidden layer of the Transformer-based Autoencoder to get common representations, then use a Migration Prediction layer and a Rating Prediction layer to get predictions for migrations and ratings, respectively, to achieve multi-task learning. Table~\ref{tab:notation} illustrates the notations along the following sections and their dimensions.
        
        \begin{table}[t]
            \caption{Notations and Descriptions}
            \label{tab:notation}
            \centering
            \scriptsize
            \setlength{\tabcolsep}{3.8mm}{
            \begin{tabular}{cll}
                \toprule
                Notations & Description & Dimension\\
                \midrule
                $T$ & Time range of historical data & Scalar\\
                $N$ & Number of samples & Scalar\\
                $D$ & Dimension of company information & Scalar\\
                $X$ & Historical information of a company & $T \times D$\\
                $\hat{X}$ & 1-year-lagged information of $X$ & $T \times D$\\
                $H$ & Positional encoded embedding for $X$ & $T \times D$\\
                $\hat{H}$ & Positional encoded embedding for $\hat{X}$ & $T \times D$\\
                $Z$ & Hidden features of Transformer-based Autoencoder & $T \times 256$\\
                $A$ & Output of Transformer-based Decoder & $T \times D$\\
                $M$ & Prediction for migration after 1 year & $T \times 3$\\
                $R$ & Prediction for rating after 1 year & $T \times 14$\\
                $Y_M$ & One-hot ground truth of migration after 12 months & $T \times 3$\\
                $Y_R$ & One-hot ground truth of rating after 12 months & $T \times 14$\\
                \bottomrule
            \end{tabular}
            }
            \vspace{-4mm}
        \end{table}

    \subsection{Objective Function}
        
        Our model consists of three parts: Positional Encoding, Transformer-based Autoencoder, and Multi-task Prediction. In this subsection, $X$ and $\hat{X}$ represent input data of the first and second (1-year-lagged) period, respectively. We use $\theta = \{\theta_E, \theta_D, \theta_P\}$ to denote the trainable parameter set of Transformer-based Encoder, Transformer-based Decoder, and Multi-task Prediction in the proposed model. Specifically, $\theta_E = \{\theta_O, \theta_Q, \theta_K, \theta_V, \theta_F\}$ denotes the trainable parameter set in the Transformer-based Encoder part, where $\theta_O$, $\theta_Q$, $\theta_K$, and $\theta_V$ are the trainable parameters of the linear layer, queries, keys, and values in the multi-head attention layer, respectively, and $\theta_F$ is the trainable parameters of the fully connected layer. $\theta_D$ denotes the trainable parameter set in the Transformer-based Decoder part, which has similar components as $\theta_E$ but different dimensions. $\theta_P = \{\theta_C, \theta_M, \theta_R\}$ denotes the trainable parameters in the Prediction part, where $\theta_C$, $\theta_M$, and $\theta_R$ are the trainable parameters of the Common Embedding, Migration Prediction, and Rating Prediction, respectively. Our goal is to minimize the objective function by adjusting $\theta$ with the model and data. Each part of the model is detailed as follows.
        
        \textbf{Positional Encoding}. Positional Encoding~\citep{vaswani2017attention} is designed for the model to make use of the order of the sequence without involving recurrence and convolution. To inject some information about the relative or absolute position of the time-series data, we adopt a fixed positional encoding method~\citep{gehring2017convolutional}, which is sine and cosine functions of different frequencies as follows:
        \begin{equation}
           PE_{(pos,2i)} = \sin(pos/10000^{2i/d}),
        \end{equation}
        \begin{equation}
           PE_{(pos,2i+1)} = \cos(pos/10000^{2i/d}),
        \end{equation}
        where $pos$ is the position, and $i$ is the dimension. That is, each dimension of the positional encoding corresponds to a sinusoid. The wavelengths form a geometric progression from $2\pi$ to $10000\cdot2\pi$. Since for any fixed offset $k$, $PE_{pos+k}$ can be represented as a linear function of $PE_{pos}$, it would allow the model to easily learn to attend by relative positions. Then the positional encoding is added to the input data as follows:
        \begin{equation}
           H = X + PE,\ \textup{and}\ 
           \hat{H} = \hat{X} + PE,
        \end{equation}
        where $H$ and $\hat{H}$ are embeddings for $X$ and $\hat{X}$, respectively. While this positional encoding method is fixed, there are no learnable parameters in this part. We do not add an additional learned embedding layer to convert input data of the first and second period to a same dimension as \citet{vaswani2017attention} did because the dimensions of the two pieces of data not only have the same length but also have the same meaning.
        
        \textbf{Transformer-based Autoencoder}. The Transformer-based Autoencoder is built by a Transformer-based Encoder~\citep{vaswani2017attention} and a Transformer-based Decoder~\citep{vaswani2017attention}, where a hidden representation $Z$ is generated.
        
        The Transformer-based Encoder has two sub-layers. The first is a multi-head self-attention mechanism, and the second is a simple, position-wise, fully connected feed-forward network. There is a residual connection~\citep{he2016deep} around each of the two sub-layers, followed by layer normalization~\citep{ba2016layer}. It first defines a scaled dot-product attention function, which is shown as:
        \begin{equation}
            \label{eq:att}
           {\rm Attention}(Q,K,V) = {\rm Softmax}(\frac{Q \cdot K^\top}{\sqrt{d_k}})V,
        \end{equation}
        where $Q$, $K$, and $V$ are embeddings generated by $H$ and $\theta_A$, denoting queries, keys, and values, respectively. Here $d_k$ is the dimension of $K$. Based on Eq.~\eqref{eq:att}, multi-head attention allows the model to jointly attend to information from different representation subspaces at different positions, which can be formulated as:
        \begin{equation}
           E = {\rm Concat}({\rm head_1}, {\rm head_2}, ..., {\rm head_h}) \theta_O,
        \end{equation}
        \begin{equation}
           {\rm head_i} = {\rm Attention}(H \cdot \theta_{Q_i},H \cdot \theta_{K_i},H \cdot \theta_{V_i}), i \in [1,h],
        \end{equation}
        where $h$ denotes the number of heads, and  $\theta_O$, $\theta_Q=\{\theta_{Q_1}, ..., \theta_{Q_h}\}$, $\theta_K=\{\theta_{K_1}, ..., \theta_{K_h}\}$, and $\theta_V=\{\theta_{V_1}, ..., \theta_{V_h}\}$ are learnable parameters. With a layer normalization function ${\rm LayerNorm(\cdot)}$~\citep{ba2016layer}, the output of the first sub-layer in Transformer-based Encoder can be written by:
        \begin{equation}
           \Tilde{E} = {\rm LayerNorm}(H+E).
        \end{equation}
        Then for the second fully connected layer together with another normalization layer, the Transformer-based Encoder generates a hidden representation $Z$ by the following equation:
        \begin{equation}
           Z = {\rm LayerNorm}(\Tilde{E}+\Tilde{E} \cdot \theta_F),
        \end{equation}
        where $\theta_F$ denotes the learnable parameters in the fully connected layer.
        
        Similar to the encoder part, the Transformer-based Decoder also contains a multi-head self-attention layer as well as a position-wise fully connected layer. To make it easy, we use $Z = Encoder(H, \theta_E)$ to denote the operations and outputs of the encoder part, and $A = Decoder(Z, \theta_D)$ the decoder part.
        
        \textbf{Multi-task Prediction}. The migration prediction task is highly related to the rating prediction task because the migrations can be inferred by ratings. To help improve the performance of migration prediction, we design a multi-task learning method, which can forecast migrations and ratings simultaneously. To achieve this, we first build a common embedding layer to generate representations for both prediction tasks, which can be written as:
        \begin{equation}
           C = {\rm ReLU}(Z \cdot \theta_C),
        \end{equation}
        where ${\rm ReLU}(x) = \max (x, 0)$ is the activation function, and $\theta_C$ denotes the learnable parameters. Then we build the migration prediction part and rating prediction part on top of the common embedding layer as follows:
        \begin{equation}
           M = {\rm Softmax}(C \cdot \theta_M),
        \end{equation}
        \begin{equation}
           R = {\rm Softmax}(C \cdot \theta_R),
        \end{equation}
        where $M$ is a 3-dimensional vector denoting the probabilities of upgraded, unchanged, and downgraded, and $R$ is a 14-dimensional vector denoting the probabilities of under different rating levels after 12 months.
        
        \textbf{Overall Objective Function}. Our objective function contains 3 parts. The first one is the mean squared error of $A$ and $\hat{H}$, driving the model to learn an envisioning ability in the Transformer-based Autoencoder part, which is:
        \begin{equation}
           \mathcal{L}_A = \frac{1}{T}\sum_t^T \frac{1}{D}\sum_i^D (A_{t,i} - \hat{H}_{t,i})^2,
           \label{eq:loss_a}
        \end{equation}
        where $T$ is the time length of historical data, and $D$ is the dimension of input features. The other two parts of the objective function are cross-entropy losses for migration prediction and rating prediction:
        \begin{equation}
           \mathcal{L}_M = \frac{1}{T} \sum_{t}^T\sum_{p}^3 (1-{Y_M}_{t,p}) M_{t,p},
           \label{eq:loss_m}
        \end{equation}
        \begin{equation}
           \mathcal{L}_R = \frac{1}{T} \sum_{t}^T\sum_{p}^{14} (1-{Y_R}_{t,p}) R_{t,p},
           \label{eq:loss_r}
        \end{equation}
        where ${Y_M}$ and ${Y_R}$ are one-hot encodings of ground truth for migration and rating, respectively.
        
        Combining Eq.\eqref{eq:loss_a}, Eq.\eqref{eq:loss_m}, and Eq.\eqref{eq:loss_r}, our overall objective function is:
        \begin{equation}
           \min_{\theta}~\mathcal{L}_A + \alpha \mathcal{L}_M + \beta \mathcal{L}_R,
           \label{eq:obj}
        \end{equation}
        where $\alpha$ and $\beta$ are hyperparameters controlling the weights of $\mathcal{L}_M$ and $\mathcal{L}_R$. We adopt Adam optimizer~\citep{kingma2014adam} to minimize the objective function. More detailed settings can be found in the following section.
    
\section{Experiments}
    In this section, we first introduce the experimental setting, then we evaluate our model. Finally, we provide some insightful experiments to demonstrate the effectiveness of the proposed model.
    
    \subsection{Experimental Setting}
        \label{sec:settings}
        In our experiments, we aim to predict rating migrations after 12 months for companies. We also set the historical data length of each data point to be exactly 12 months to meet the prediction time range. The total test period is from 2005/01/01 to 2020/12/31. We adopt an expanding window for the training set and re-calibrate the models every 3 months. For example, we train the model with data from 1997/01/01 to 2004/12/31 (available migration labels are from 1997/01/01 to 2003/12/31) and use the trained model to predict the 12-month-later migration of a corporation with its information data on and before 2005/01/01. Here the ground truth of 12-month-later migration is decided on 2006/01/01.
        
        
        \textbf{Baseline methods}. We choose 6 classical classification models as baseline methods for comparison: K-NearestNeighbor (KNN)~\citep{fix1989discriminatory}, Support Vector Machine (SVM)~\citep{boser1992training}, Random Forest (RF)~\citep{breiman2001random}, Multi-Layer Perceptron (MLP)~\citep{he2015delving}, AdaBoost~\citep{hastie2009multi}, and Gaussian Naïve Bayes (NB)~\citep{tan2016introduction}. These methods are implemented based on Sklearn.\footnote{\href{https://scikit-learn.org/stable/}{https://scikit-learn.org/stable/}} Among them, SVM, MLP, and ensemble methods like AdaBoost are adopted by previous studies~\cite{lee2007application,kim2012ensemble,jabeur2020machine} on rating prediction. Beyond these, we also choose 5 time-series prediction models for comparison: Long Short-Term Memory (LSTM)~\citep{hochreiter1997long}, DeepAR~\cite{salinas2020deepar}, Transformer~\cite{vaswani2017attention}, LogTrans~\cite{li2019enhancing}, and Informer~\cite{zhou2021informer}. LSTM is implemented based on Keras,\footnote{\href{https://keras.io/}{https://keras.io/}} Transformer is implemented based on Pytorch.\footnote{\href{https://pytorch.org/docs/stable/generated/torch.nn.Transformer.html}{https://pytorch.org/docs/stable/generated/torch.nn.Transformer.html}} For the rest of methods, we use the public source codes provided by the authors on Github.\footnote{\href{https://github.com/}{https://github.com/}}
        

        \textbf{Implementation details}. For the proposed Multi-task Envisioning Transformer-based Autoencoder (META), we implement it by Pytorch~\citep{paszke2019pytorch}, adopt Adam optimizer~\citep{kingma2014adam} with a learning rate of 0.0001, read the data with a batch size of 1024, and run 3000 epochs for each re-calibration to train the model. The values of $\alpha$ and $\beta$ in the objective function are set to 1 by default.
        
            \begin{figure*}[t]
        \centering
        \subfigure[$F_1$-Up by year]{
                \begin{minipage}[t]{0.31\textwidth}
                \centering
                \label{fig:f1_year_up}
                \includegraphics[scale=0.15]{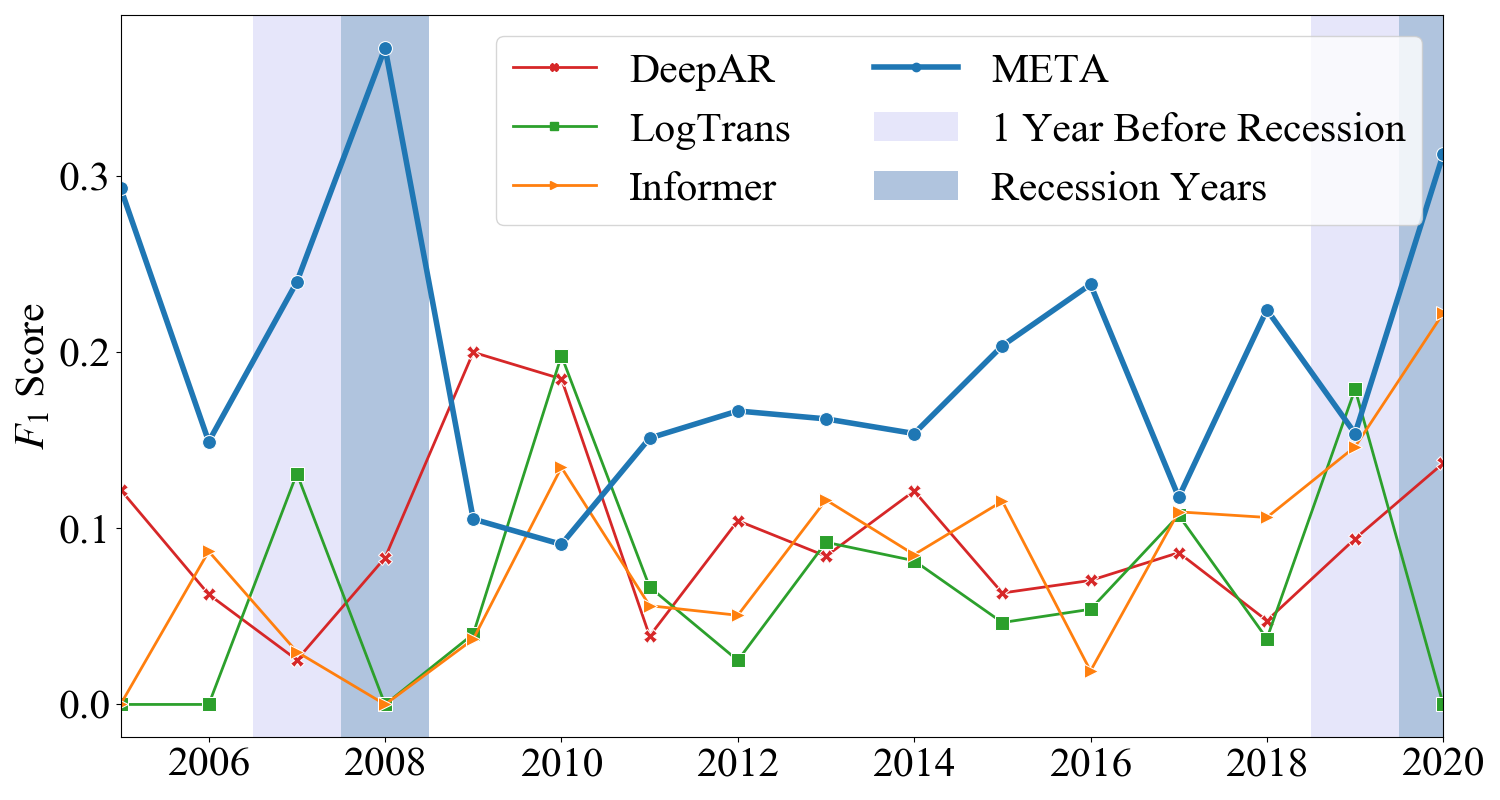}
                \end{minipage}
            }
        \subfigure[$F_1$-Down by year]{
                \begin{minipage}[t]{0.31\textwidth}
                \centering
                \label{fig:f1_year_down}
                \includegraphics[scale=0.15]{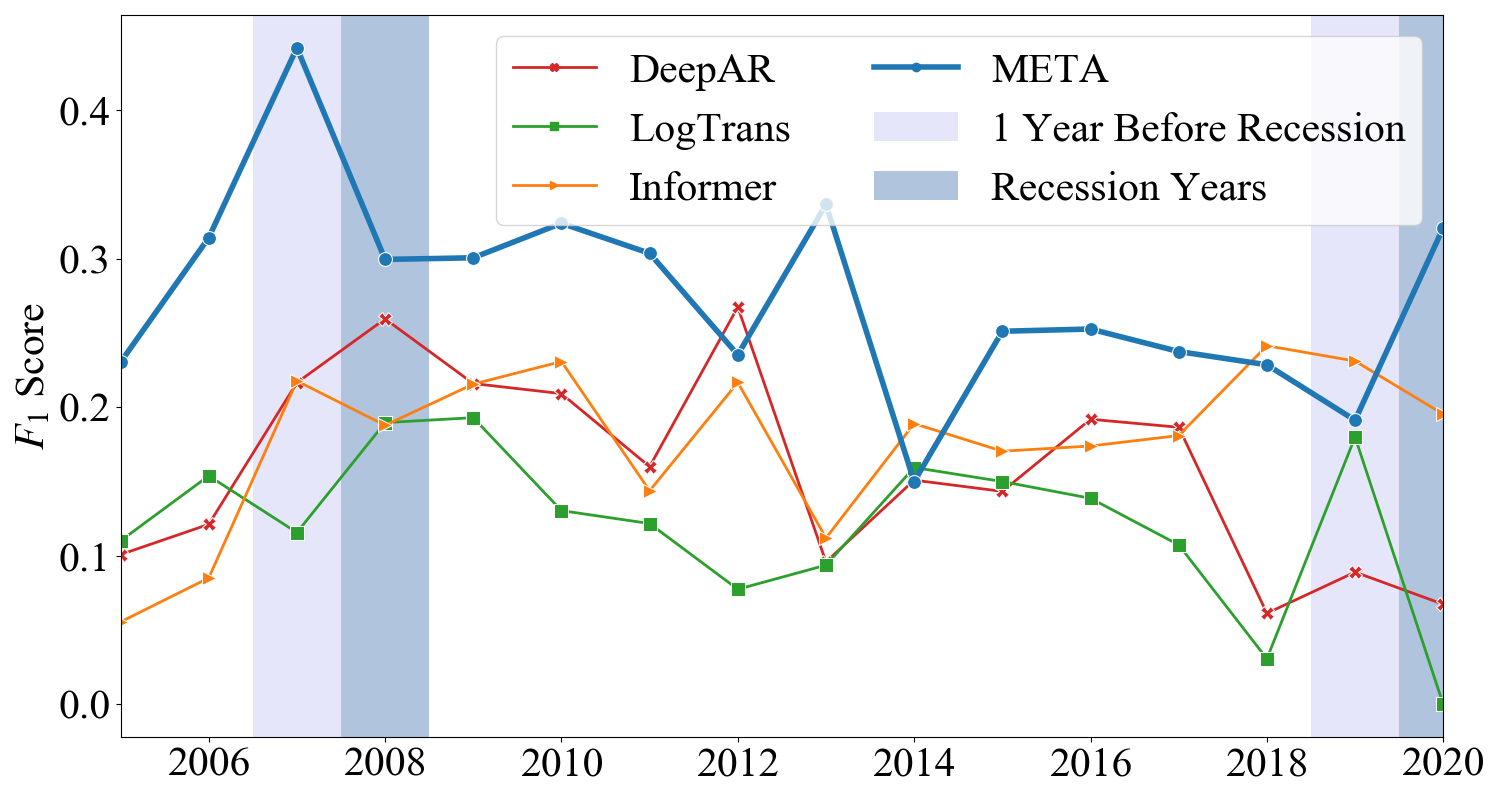}
                \end{minipage}
            }
        \subfigure[$F_1$-Down for companies of different ratings]{
                \begin{minipage}[t]{0.31\textwidth}
                \centering
                \label{fig:f1_ratings}
                \includegraphics[scale=0.15]{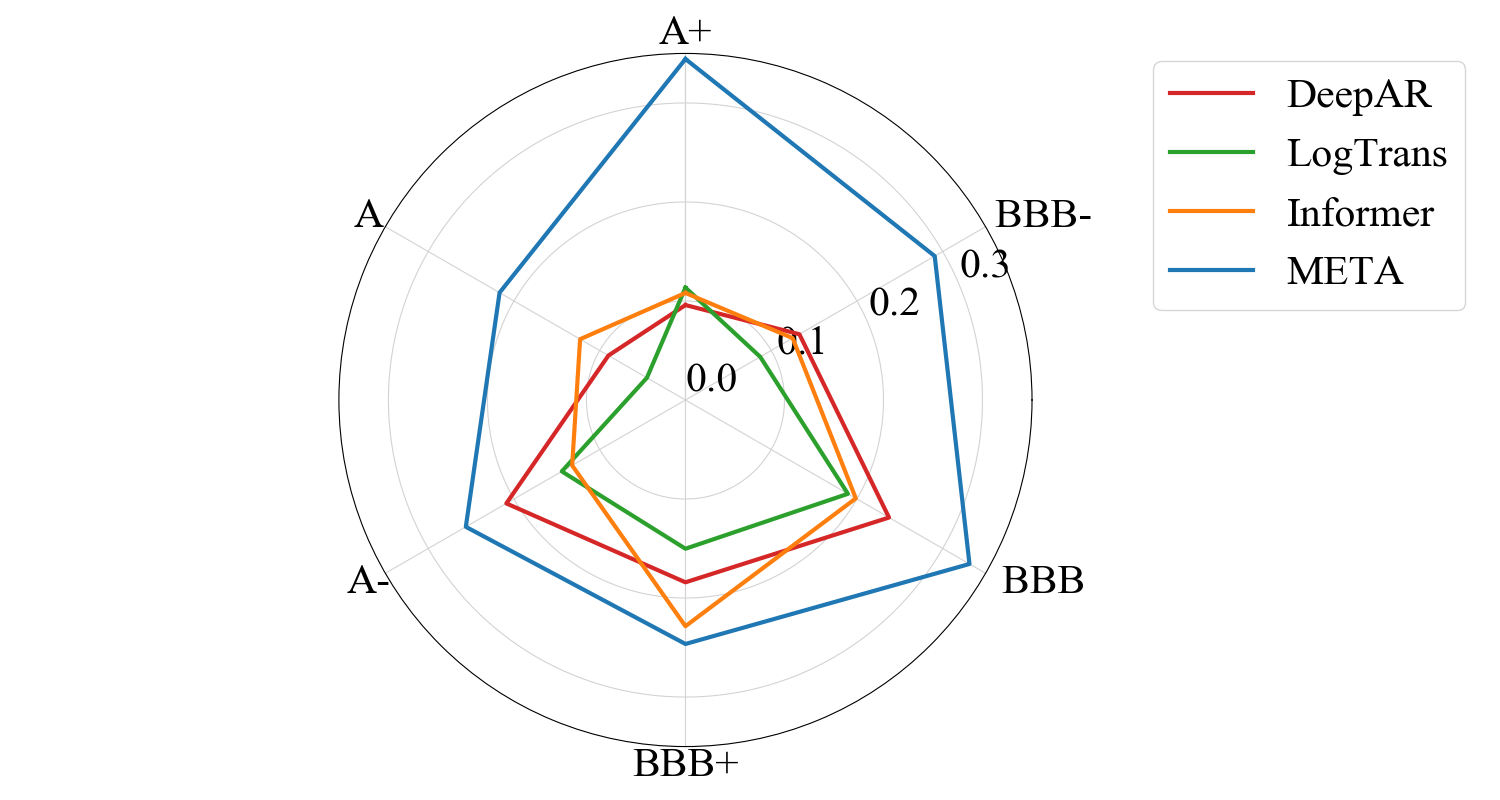}
                \end{minipage}
            }
        
        \vspace{-4mm}
        \caption{Early prediction performance of four time-series algorithms. (a-b) $F_1$-Up and $F_1$-Down by year in the test period from 2005 to 2022, where the years in recessions and early-recessions are highlighted with light blue and dark blue shadow; (c) Radar plot of four time-series algorithms for $F_1$-Down in 6 rating groups, where most rating migrations occur.}
        \label{fig:pred_res}
        \vspace{-2mm}
    \end{figure*}
        
        \textbf{Evaluation metric}. While the numbers of upgraded, downgraded, and unchanged companies are imbalanced as shown in Figure~\ref{fig:stat_up_down}, we decide to use $F_1$ score for both upgrades and downgrades as our evaluation metric. In supplement to $F_1$ score, we also provide accuracy to verify that the models are not performing a poor prediction for unchanged situations. The calculations of $F_1$ score and accuracy are as follows:
        \begin{equation}
           {\rm precision} = \frac{tp}{tp + fp},\ \ \ \ {\rm recall} = \frac{tp}{tp + fn}
        \end{equation}
        \begin{equation}
           F_1 = \frac{2*{\rm recall}*{\rm precision}}{{\rm recall} + {\rm precision}},
        \end{equation}
        \begin{equation}
           {\rm Accuracy} = \frac{tp+tn}{tp+fp+fn+tn},
        \end{equation}
        where $tp$, $fp$, $fn$, $tn$ are the numbers of true positives, false positives, false negatives, true negatives, respectively. Here we take $F_1$ score for upgrades as an example, then $tp$ denotes the number of upgraded samples that are predicted correctly by the model, $fp$ denotes the number of downgraded or unchanged samples that are predicted incorrectly by the model, $fn$ denotes the number of upgraded samples that are predicted incorrectly by the model, and $tn$ is the number of downgraded or unchanged samples that are predicted correctly by the model.
        
        
        Since the early rating migration problem is an extremely imbalanced prediction problem, $F_1$ is more suitable than Accuracy for algorithmic evaluation. Moreover, in this financial application, we care more about the downgrade migration, i.e., $F_1$-Down, because of its strong implication for not only investment professionals but also risk managers and regulators. 
    
    \subsection{Algorithmic Performance}

        Based on the settings mentioned above, we run all the methods and gather their predictions from 2005/01/01 to 2020/12/31. Table~\ref{tab:perform} shows the overall performance of all models during the test period. Among classical non-time-series baseline models, SVM achieves the highest accuracy but gets very low $F_1$-Up and $F_1$-Down. Furthermore,  SVM gives a very high number of unchanged predictions, which happens to achieve the best accuracy on this unbalanced dataset where 85.77\% samples are unchanged. This is also a good example to illustrate why we choose $F_1$-Up and $F_1$-Down for evaluation. Another interesting result is achieved by NB, whose accuracy is low at 21.60\%, indicating that NB is trying not to predict unchanged situations for samples, and the $F_1$-Up and $F_1$-Down are somehow good because it achieves a high recall. The time-series algorithms, including two classical deep models and three state-of-the-art models, deliver similar performance as non-time-series methods. The reason is that there is a 1-year-gap between their training and test data, and the data distribution may change during this period. On the contrary, META leverages the influence of this gap by adopting positional encoding and autoencoder to use the data during the gap period. Therefore, the representation learned by META is capable of containing future trends for early prediction. It not only handles the time-series data but also has the envisioning ability, which makes full use of all data in the training set no matter whether the data is labeled or not. Moreover, different from baseline methods, META is the only model that performs multi-task learning, which also helps improve the model in providing stable and reliable predictions. Compared with baseline methods, our proposed META outperforms and has a significant advantage over the baseline and state-of-the-art methods on both $F_1$-Up and $F_1$-Down.

        \begin{table}[t]
            \centering
            \caption{Algorithmic performance of 6 non-time-series and 6 time-series methods on 12-month early rating migration prediction from 2005 to 2020 by $F_1$-Up, $F_1$-Down and Accuracy}\vspace{-2mm}
            \scriptsize
            \setlength{\tabcolsep}{5mm}{
            \begin{tabular}{l|ccc}
            \toprule
            Methods    & $F_1$-Up    & $F_1$-Down  & Accuracy  \\
            \midrule
            AdaBoost & 0.0432 & 0.0967 & 0.8074  \\
            KNN      & 0.0964 & 0.1535 & 0.7392  \\
            MLP      & 0.0803 & 0.1744 & 0.7319  \\
            NB       & 0.1422 & 0.2076 & 0.2160  \\
            RF       & 0.1368 & 0.2312 & 0.7347  \\
            SVM      & 0.0000 & 0.0085 & \textbf{0.8242}  \\
            \midrule
            LSTM     & 0.0590 & 0.1369 & 0.7324  \\
            Transformer & 0.0586 & 0.1366 & 0.6982 \\
            DeepAR  & 0.0865 & 0.1698 & 0.7322 \\
            DeepAR* & 0.0925 & 0.1044 & 0.4234 \\
            LogTrans & 0.0755 & 0.1329 & 0.7759\\
            LogTrans* & 0.0804 & 0.0864 & 0.4503\\
            Informer & 0.0886 & 0.1848 & 0.7659\\
            Informer* & 0.0833 & 0.0935 & 0.4540\\
            \textbf{META} (Ours)     & \textbf{0.1864} & \textbf{0.2738} & 0.7909  \\
            \bottomrule
            \end{tabular}}
            \begin{tablenotes}
            \item Note: ``*'' denotes to use the method to predict the ratings first, and then the migration is derived from the predicted ratings.
        \end{tablenotes}
            \label{tab:perform}\vspace{-4mm}
        \end{table}
        
        Figure~\ref{fig:pred_res} shows the detailed early prediction performance of three state-of-the-art time-series algorithms and META by year and by rating group, where the years in recessions and early-recessions are highlighted with light blue and dark blue shadow, and five rating groups with frequent migrations are chosen. META outperforms other competitive methods by large margins in most years, which demonstrates the effectiveness of our proposed method for early prediction. Another noteworthy observation is that META performs even better in predicting downgrade risk in the time periods right before and around recessions. In financial applications, a higher early predictive capability is particularly beneficial right before major economic and market downturns. To this end, META performs much better in the recession and early-recession periods than in normal periods, which can be regarded as a significant advantage in practical use. Moreover, if we take a close look at the rating groups (A+, A, A-, BBB+, BBB, and BBB-), where rating migrations frequently occur and are of great value in financial investigation, our META also outperforms other methods as well.

    \subsection{In-depth Exploration}
    Beyond the above algorithmic comparisons with several baseline methods, we also provide in-depth explorations of our proposed META in terms of multi-task prediction, envisioning verification, and different gap periods for early prediction.

            \textbf{Multi-task Prediction}. The proposed META is a multi-task model, and it generates two kinds of predictions: migration and rating. While inferring from the rating prediction result, we can easily get another migration result that may not be the same as the migration prediction by META directly. Table~\ref{tab:task} shows our exploration of META in performing different tasks and applying different results. When predicting ratings and changing them to migrations, the accuracy is lower than directly predicting migration. It is because there are 14 levels of ratings while only 3 kinds of migrations, which brings difficulty in predicting unchanged situations. While only performing the migration prediction task, the model cannot learn from training data properly due to unbalanced and noisy data. When performing both tasks together, the migration prediction result is significantly improved with the guidance of the rating prediction task.

             \begin{table}[t]
                \centering
                \caption{Performance of META under different tasks. Check-mark denotes that the corresponding task is performed. Results from rating to migration mean that the migration is calculated based on the rating prediction results of META, and the others are directly predicted by META}\vspace{-2mm}
                \scriptsize
                \setlength{\tabcolsep}{1.5mm}{
                \begin{tabular}{cccccc}
                \toprule
                Migration  & Rating     & Results From & $F_1$-Up & $F_1$-Down & Accuracy  \\
                \midrule
                           & \checkmark & Rating $\to$ Migration      & \textbf{0.1944} & 0.1211   & 0.5874    \\
                \checkmark &            & Directly Predicting Migration    & 0.0577 & 0.1203   & 0.7724    \\
                \checkmark & \checkmark & Rating $\to$ Migration      & 0.1278 & 0.1561   & 0.6088    \\
                \checkmark & \checkmark & Multi-task Predicting Migration    & 0.1864 & \textbf{0.2738}   & 0.7909    \\
                \bottomrule
                \end{tabular}}
                \label{tab:task}\vspace{-4mm}
            \end{table}

            \begin{table}[t]
                \centering
                \caption{Algorithmic performance of 6 non-time-series and 6 time-series methods on no gap rating migration prediction from 2005 to 2020 by $F_1$-Up, $F_1$-Down and Accuracy}\vspace{-2mm}
                \scriptsize
                \setlength{\tabcolsep}{5mm}{
                \begin{tabular}{l|ccc}
                \toprule
                         & $F_1$-Up & $F_1$-Down & Accuracy \\
                \midrule
                Adaboost & 0.0525 & 0.1112   & 0.8164\\
                KNN      & 0.1752 & 0.2285   & 0.7693\\
                MLP      & 0.1429 & 0.2317   & 0.7554\\
                NB       & 0.1354 & 0.2018   & 0.2155\\
                RF       & 0.1644 & 0.2553   & 0.7116\\
                SVM      & 0.0000 & 0.0068   & 0.8332\\
                \midrule
                LSTM     & 0.0345 & 0.0902   & 0.8559\\
                Transformer & 0.1410 & 0.1748 & 0.7149 \\
                DeepAR & 0.1221 & 0.2210 & 0.7486 \\
                LogTrans & 0.1382 & 0.2083 & 0.7556\\
                Informer & 0.1716 & 0.2415 & 0.7818\\
                \textbf{META} (no gap) & \textbf{0.2174} & \textbf{0.2812}   & 0.7966\\
                \midrule
                \textbf{META} (12-month gap)     & \textbf{0.1864} & \textbf{0.2738} & 0.7909\\
                \bottomrule
                \end{tabular}}
                \label{tab:gap}
                 \vspace{-4mm}
            \end{table}
            
            \textbf{Envisioning Verification}. To demonstrate the envisioning ability of META, we set a pseudo environment assuming that we always know the migration results for all the past days, which would eliminate the 1-year gap between training and test set. Table~\ref{tab:gap} shows the results of META under the pseudo setting. META under the pseudo setting achieves a better result than with the normal setting, but not significant, especially on $F_1$-Down, the improvement of which is less than 1\%. We also run all baseline models under the pseudo setting as listed in Table~\ref{tab:gap}, from where we can see that all models have a better performance under this setting. It is noticeable that META can still outperform all baseline models even when META uses the normal setting and others are with the pseudo setting, which demonstrates the effectiveness of META.

\begin{figure}[t]
            \centering
            \subfigure[$F_1$-Up]{
                    \begin{minipage}[t]{0.222\textwidth}
                    \centering
                    \label{fig:f1_3_6_up}
                    \includegraphics[scale=0.15]{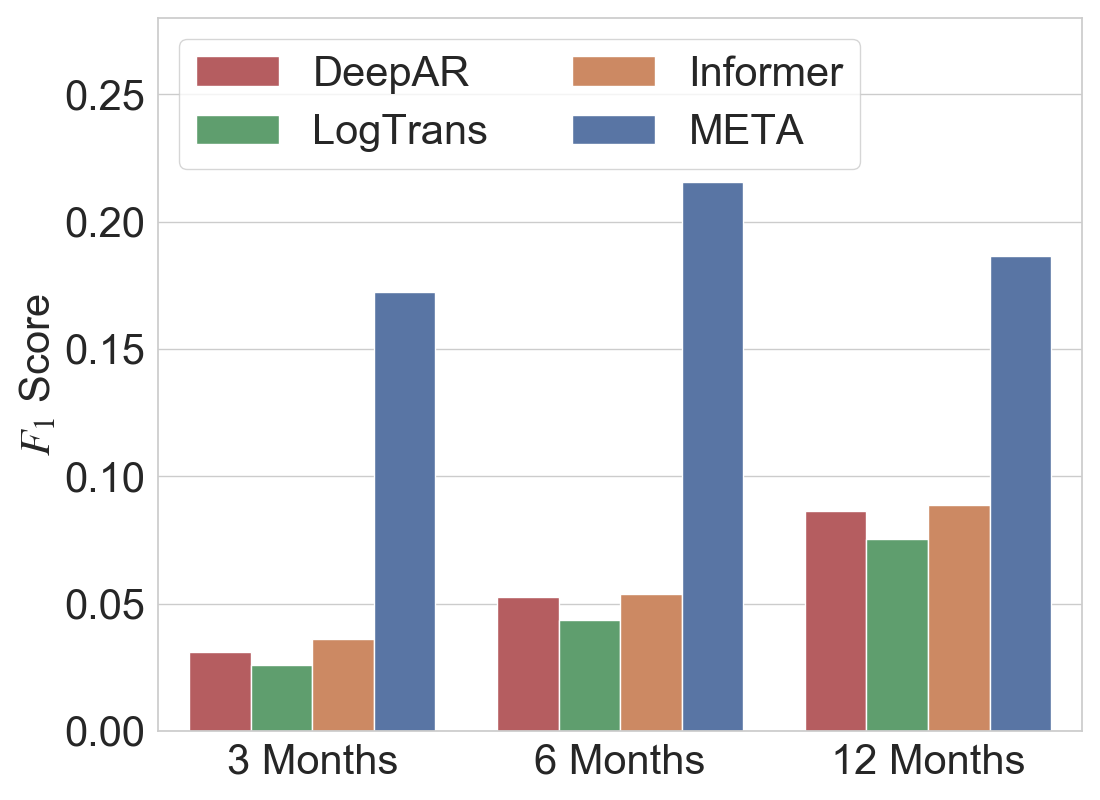}
                    \end{minipage}
                }
            \subfigure[$F_1$-Down]{
                    \begin{minipage}[t]{0.222\textwidth}
                    \centering
                    \label{fig:f1_3_6_down}
                    \includegraphics[scale=0.15]{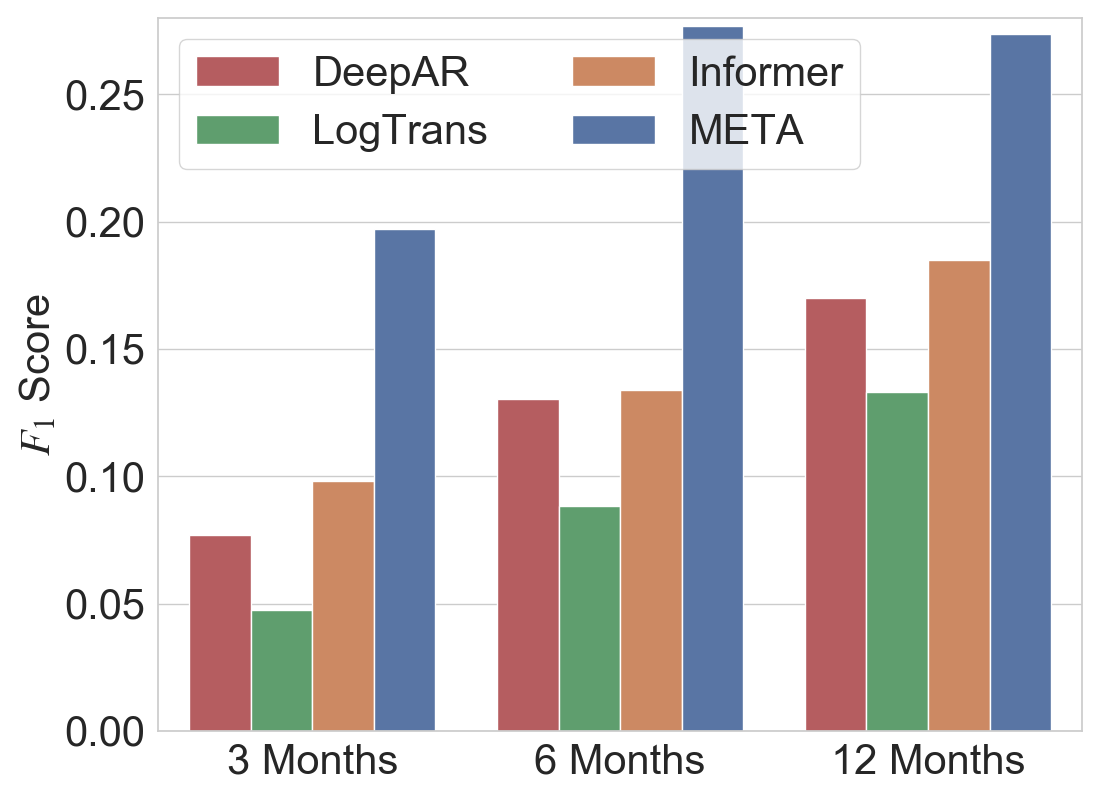}
                    \end{minipage}
                }
            \vspace{-4mm}
            \caption{Early migration prediction of four time-series algorithms with different gap periods.}
            \label{fig:pred_time}\vspace{-4mm}
            \vspace{-2mm}
        \end{figure}

        
        \textbf{Early Prediction Performance with Different Gap Periods}. So far, we use a 12-month early prediction setting. Here we demonstrate the performance of our META with different gap periods. Figure~\ref{fig:pred_time} shows the early migration prediction of three state-of-the-art time-series algorithms and META with 3-month, 6-month, and 12-month gap periods, respectively. In both upgrade and downgrade scenarios, META outperforms other competitive methods in all the cases, which demonstrates the envisioning capacity of META in different gap periods. 

\section{Conclusion}
    In this paper, we focus on the rating migration prediction problem. With historical information on companies, we aim to predict how the rating of these companies will change (upgraded, downgraded, and unchanged) 12 months later. We propose Multi-task Envisioning Transformer-based Autoencoder (META) to address this problem. Specifically, we first adopt a Positional Encoding layer to incorporate time information with company information, then we design a Transformer-based Autoencoder to learn envisioning ability from historical data. Finally, we use Multi-task Prediction to get predictions for both migrations and ratings simultaneously. Experimental results show that META outperforms all 11 baseline methods, which demonstrates the effectiveness of our proposed method. Our proposed model also has the advantage of higher precision during time periods that are most relevant for its applications, i.e., before and during economic recessions.


\bibliographystyle{ACM-Reference-Format}
\bibliography{sample-base}


\begin{thebibliography}{63}


\ifx \showCODEN    \undefined \def \showCODEN     #1{\unskip}     \fi
\ifx \showDOI      \undefined \def \showDOI       #1{#1}\fi
\ifx \showISBNx    \undefined \def \showISBNx     #1{\unskip}     \fi
\ifx \showISBNxiii \undefined \def \showISBNxiii  #1{\unskip}     \fi
\ifx \showISSN     \undefined \def \showISSN      #1{\unskip}     \fi
\ifx \showLCCN     \undefined \def \showLCCN      #1{\unskip}     \fi
\ifx \shownote     \undefined \def \shownote      #1{#1}          \fi
\ifx \showarticletitle \undefined \def \showarticletitle #1{#1}   \fi
\ifx \showURL      \undefined \def \showURL       {\relax}        \fi
\providecommand\bibfield[2]{#2}
\providecommand\bibinfo[2]{#2}
\providecommand\natexlab[1]{#1}
\providecommand\showeprint[2][]{arXiv:#2}

\bibitem[\protect\citeauthoryear{Ahelegbey, Giudici, and
  Hadji-Misheva}{Ahelegbey et~al\mbox{.}}{2019}]%
        {ahelegbey2019latent}
\bibfield{author}{\bibinfo{person}{Daniel~Felix Ahelegbey},
  \bibinfo{person}{Paolo Giudici}, {and} \bibinfo{person}{Branka
  Hadji-Misheva}.} \bibinfo{year}{2019}\natexlab{}.
\newblock \showarticletitle{Latent factor models for credit scoring in P2P
  systems}.
\newblock \bibinfo{journal}{\emph{Physica A: Statistical Mechanics and its
  Applications}}  \bibinfo{volume}{522} (\bibinfo{year}{2019}),
  \bibinfo{pages}{112--121}.
\newblock


\bibitem[\protect\citeauthoryear{Altman}{Altman}{1998}]%
        {altman1998importance}
\bibfield{author}{\bibinfo{person}{Edward~I Altman}.}
  \bibinfo{year}{1998}\natexlab{}.
\newblock \showarticletitle{The importance and subtlety of credit rating
  migration}.
\newblock \bibinfo{journal}{\emph{Journal of Banking \& Finance}}
  \bibinfo{volume}{22}, \bibinfo{number}{10-11} (\bibinfo{year}{1998}),
  \bibinfo{pages}{1231--1247}.
\newblock


\bibitem[\protect\citeauthoryear{Ba, Kiros, and Hinton}{Ba
  et~al\mbox{.}}{2016}]%
        {ba2016layer}
\bibfield{author}{\bibinfo{person}{Jimmy~Lei Ba}, \bibinfo{person}{Jamie~Ryan
  Kiros}, {and} \bibinfo{person}{Geoffrey~E Hinton}.}
  \bibinfo{year}{2016}\natexlab{}.
\newblock \showarticletitle{Layer normalization}.
\newblock \bibinfo{journal}{\emph{arXiv preprint arXiv:1607.06450}}
  (\bibinfo{year}{2016}).
\newblock


\bibitem[\protect\citeauthoryear{Bai, Kolter, and Koltun}{Bai
  et~al\mbox{.}}{2018}]%
        {bai2018empirical}
\bibfield{author}{\bibinfo{person}{Shaojie Bai}, \bibinfo{person}{J~Zico
  Kolter}, {and} \bibinfo{person}{Vladlen Koltun}.}
  \bibinfo{year}{2018}\natexlab{}.
\newblock \showarticletitle{An empirical evaluation of generic convolutional
  and recurrent networks for sequence modeling}.
\newblock \bibinfo{journal}{\emph{arXiv preprint arXiv:1803.01271}}
  (\bibinfo{year}{2018}).
\newblock


\bibitem[\protect\citeauthoryear{Boser, Guyon, and Vapnik}{Boser
  et~al\mbox{.}}{1992}]%
        {boser1992training}
\bibfield{author}{\bibinfo{person}{Bernhard~E Boser},
  \bibinfo{person}{Isabelle~M Guyon}, {and} \bibinfo{person}{Vladimir~N
  Vapnik}.} \bibinfo{year}{1992}\natexlab{}.
\newblock \showarticletitle{A training algorithm for optimal margin
  classifiers}. In \bibinfo{booktitle}{\emph{Annual Workshop on Computational
  Learning Theory}}.
\newblock


\bibitem[\protect\citeauthoryear{Breiman}{Breiman}{2001}]%
        {breiman2001random}
\bibfield{author}{\bibinfo{person}{Leo Breiman}.}
  \bibinfo{year}{2001}\natexlab{}.
\newblock \showarticletitle{Random forests}.
\newblock \bibinfo{journal}{\emph{Machine learning}} \bibinfo{volume}{45},
  \bibinfo{number}{1} (\bibinfo{year}{2001}), \bibinfo{pages}{5--32}.
\newblock


\bibitem[\protect\citeauthoryear{Cao, Yang, and Yu}{Cao et~al\mbox{.}}{2021}]%
        {cao2021data}
\bibfield{author}{\bibinfo{person}{Longbing Cao}, \bibinfo{person}{Qiang Yang},
  {and} \bibinfo{person}{Philip~S Yu}.} \bibinfo{year}{2021}\natexlab{}.
\newblock \showarticletitle{Data science and AI in FinTech: An overview}.
\newblock \bibinfo{journal}{\emph{International Journal of Data Science and
  Analytics}} \bibinfo{volume}{12}, \bibinfo{number}{2} (\bibinfo{year}{2021}),
  \bibinfo{pages}{81--99}.
\newblock


\bibitem[\protect\citeauthoryear{Chauchat, Rakotomalala, Carloz, and
  Pelletier}{Chauchat et~al\mbox{.}}{2001}]%
        {chauchat2001targeting}
\bibfield{author}{\bibinfo{person}{JH Chauchat}, \bibinfo{person}{Ricco
  Rakotomalala}, \bibinfo{person}{M Carloz}, {and} \bibinfo{person}{C
  Pelletier}.} \bibinfo{year}{2001}\natexlab{}.
\newblock \showarticletitle{Targeting customer groups using gain and cost
  matrix; a marketing application}. In \bibinfo{booktitle}{\emph{Data Mining
  for Marketing Applications}}.
\newblock


\bibitem[\protect\citeauthoryear{Chen and Cui}{Chen and Cui}{2020}]%
        {chen2020utilizing}
\bibfield{author}{\bibinfo{person}{Fu Chen} {and} \bibinfo{person}{Ying Cui}.}
  \bibinfo{year}{2020}\natexlab{}.
\newblock \showarticletitle{Utilizing Student Time Series Behaviour in Learning
  Management Systems for Early Prediction of Course Performance.}
\newblock \bibinfo{journal}{\emph{Journal of Learning Analytics}}
  \bibinfo{volume}{7}, \bibinfo{number}{2} (\bibinfo{year}{2020}),
  \bibinfo{pages}{1--17}.
\newblock


\bibitem[\protect\citeauthoryear{Chen and Tsai}{Chen and Tsai}{2020}]%
        {chen2020encoding}
\bibfield{author}{\bibinfo{person}{Jun-Hao Chen} {and}
  \bibinfo{person}{Yun-Cheng Tsai}.} \bibinfo{year}{2020}\natexlab{}.
\newblock \showarticletitle{Encoding candlesticks as images for pattern
  classification using convolutional neural networks}.
\newblock \bibinfo{journal}{\emph{Financial Innovation}}  \bibinfo{volume}{6}
  (\bibinfo{year}{2020}), \bibinfo{pages}{1--19}.
\newblock


\bibitem[\protect\citeauthoryear{Chen, Wu, and Bu}{Chen et~al\mbox{.}}{2018}]%
        {chen2018stock}
\bibfield{author}{\bibinfo{person}{Yuzhou Chen}, \bibinfo{person}{Junji Wu},
  {and} \bibinfo{person}{Hui Bu}.} \bibinfo{year}{2018}\natexlab{}.
\newblock \showarticletitle{Stock market embedding and prediction: A deep
  learning method}. In \bibinfo{booktitle}{\emph{International Conference on
  Service Systems and Service Management}}.
\newblock


\bibitem[\protect\citeauthoryear{Cho, Van~Merri{\"e}nboer, Gulcehre, Bahdanau,
  Bougares, Schwenk, and Bengio}{Cho et~al\mbox{.}}{2014}]%
        {cho2014learning}
\bibfield{author}{\bibinfo{person}{Kyunghyun Cho}, \bibinfo{person}{Bart
  Van~Merri{\"e}nboer}, \bibinfo{person}{Caglar Gulcehre},
  \bibinfo{person}{Dzmitry Bahdanau}, \bibinfo{person}{Fethi Bougares},
  \bibinfo{person}{Holger Schwenk}, {and} \bibinfo{person}{Yoshua Bengio}.}
  \bibinfo{year}{2014}\natexlab{}.
\newblock \showarticletitle{Learning phrase representations using RNN
  encoder-decoder for statistical machine translation}.
\newblock \bibinfo{journal}{\emph{arXiv preprint arXiv:1406.1078}}
  (\bibinfo{year}{2014}).
\newblock


\bibitem[\protect\citeauthoryear{Das, Behera, Rath, et~al\mbox{.}}{Das
  et~al\mbox{.}}{2018}]%
        {das2018real}
\bibfield{author}{\bibinfo{person}{Sushree Das}, \bibinfo{person}{Ranjan~Kumar
  Behera}, \bibinfo{person}{Santanu~Kumar Rath}, {et~al\mbox{.}}}
  \bibinfo{year}{2018}\natexlab{}.
\newblock \showarticletitle{Real-time sentiment analysis of twitter streaming
  data for stock prediction}.
\newblock \bibinfo{journal}{\emph{Procedia Computer Science}}
  \bibinfo{volume}{132} (\bibinfo{year}{2018}), \bibinfo{pages}{956--964}.
\newblock


\bibitem[\protect\citeauthoryear{Dingli and Fournier}{Dingli and
  Fournier}{2017}]%
        {dingli2017financial}
\bibfield{author}{\bibinfo{person}{Alexiei Dingli} {and}
  \bibinfo{person}{Karl~Sant Fournier}.} \bibinfo{year}{2017}\natexlab{}.
\newblock \showarticletitle{Financial time series forecasting--a deep learning
  approach}.
\newblock \bibinfo{journal}{\emph{International Journal of Machine Learning and
  Computing}} \bibinfo{volume}{7}, \bibinfo{number}{5} (\bibinfo{year}{2017}),
  \bibinfo{pages}{118--122}.
\newblock


\bibitem[\protect\citeauthoryear{Doering, Fairbank, and Markose}{Doering
  et~al\mbox{.}}{2017}]%
        {doering2017convolutional}
\bibfield{author}{\bibinfo{person}{Jonathan Doering}, \bibinfo{person}{Michael
  Fairbank}, {and} \bibinfo{person}{Sheri Markose}.}
  \bibinfo{year}{2017}\natexlab{}.
\newblock \showarticletitle{Convolutional neural networks applied to
  high-frequency market microstructure forecasting}. In
  \bibinfo{booktitle}{\emph{Computer Science and Electronic Engineering}}.
\newblock


\bibitem[\protect\citeauthoryear{Fan, Zhang, Pan, Li, Zhang, Yuan, Wu, Wang,
  Pei, and Huang}{Fan et~al\mbox{.}}{2019}]%
        {fan2019multi}
\bibfield{author}{\bibinfo{person}{Chenyou Fan}, \bibinfo{person}{Yuze Zhang},
  \bibinfo{person}{Yi Pan}, \bibinfo{person}{Xiaoyue Li}, \bibinfo{person}{Chi
  Zhang}, \bibinfo{person}{Rong Yuan}, \bibinfo{person}{Di Wu},
  \bibinfo{person}{Wensheng Wang}, \bibinfo{person}{Jian Pei}, {and}
  \bibinfo{person}{Heng Huang}.} \bibinfo{year}{2019}\natexlab{}.
\newblock \showarticletitle{Multi-horizon time series forecasting with temporal
  attention learning}. In \bibinfo{booktitle}{\emph{ACM SIGKDD International
  Conference on Knowledge Discovery \& Data Mining}}.
\newblock


\bibitem[\protect\citeauthoryear{Fix and Hodges}{Fix and Hodges}{1989}]%
        {fix1989discriminatory}
\bibfield{author}{\bibinfo{person}{Evelyn Fix} {and}
  \bibinfo{person}{Joseph~Lawson Hodges}.} \bibinfo{year}{1989}\natexlab{}.
\newblock \showarticletitle{Discriminatory analysis. Nonparametric
  discrimination: Consistency properties}.
\newblock \bibinfo{journal}{\emph{International Statistical Review/Revue
  Internationale de Statistique}} \bibinfo{volume}{57}, \bibinfo{number}{3}
  (\bibinfo{year}{1989}), \bibinfo{pages}{238--247}.
\newblock


\bibitem[\protect\citeauthoryear{Gehring, Auli, Grangier, Yarats, and
  Dauphin}{Gehring et~al\mbox{.}}{2017}]%
        {gehring2017convolutional}
\bibfield{author}{\bibinfo{person}{Jonas Gehring}, \bibinfo{person}{Michael
  Auli}, \bibinfo{person}{David Grangier}, \bibinfo{person}{Denis Yarats},
  {and} \bibinfo{person}{Yann~N Dauphin}.} \bibinfo{year}{2017}\natexlab{}.
\newblock \showarticletitle{Convolutional sequence to sequence learning}. In
  \bibinfo{booktitle}{\emph{International Conference on Machine Learning}}.
\newblock


\bibitem[\protect\citeauthoryear{Guo and Li}{Guo and Li}{2008}]%
        {guo2008neural}
\bibfield{author}{\bibinfo{person}{Tao Guo} {and} \bibinfo{person}{Gui-Yang
  Li}.} \bibinfo{year}{2008}\natexlab{}.
\newblock \showarticletitle{Neural data mining for credit card fraud
  detection}. In \bibinfo{booktitle}{\emph{International Conference on Machine
  Learning and Cybernetics}}.
\newblock


\bibitem[\protect\citeauthoryear{H{\"a}rdle, Chen, and Overbeck}{H{\"a}rdle
  et~al\mbox{.}}{2017}]%
        {hardle2017applied}
\bibfield{author}{\bibinfo{person}{Wolfgang~Karl H{\"a}rdle},
  \bibinfo{person}{Cathy Yi-Hsuan Chen}, {and} \bibinfo{person}{Ludger
  Overbeck}.} \bibinfo{year}{2017}\natexlab{}.
\newblock \bibinfo{booktitle}{\emph{Applied quantitative finance}}.
\newblock \bibinfo{publisher}{Springer}.
\newblock


\bibitem[\protect\citeauthoryear{Hastie, Rosset, Zhu, and Zou}{Hastie
  et~al\mbox{.}}{2009}]%
        {hastie2009multi}
\bibfield{author}{\bibinfo{person}{Trevor Hastie}, \bibinfo{person}{Saharon
  Rosset}, \bibinfo{person}{Ji Zhu}, {and} \bibinfo{person}{Hui Zou}.}
  \bibinfo{year}{2009}\natexlab{}.
\newblock \showarticletitle{Multi-class adaboost}.
\newblock \bibinfo{journal}{\emph{Statistics and its Interface}}
  \bibinfo{volume}{2}, \bibinfo{number}{3} (\bibinfo{year}{2009}),
  \bibinfo{pages}{349--360}.
\newblock


\bibitem[\protect\citeauthoryear{He, Zhang, Ren, and Sun}{He
  et~al\mbox{.}}{2015}]%
        {he2015delving}
\bibfield{author}{\bibinfo{person}{Kaiming He}, \bibinfo{person}{Xiangyu
  Zhang}, \bibinfo{person}{Shaoqing Ren}, {and} \bibinfo{person}{Jian Sun}.}
  \bibinfo{year}{2015}\natexlab{}.
\newblock \showarticletitle{Delving deep into rectifiers: Surpassing
  human-level performance on imagenet classification}. In
  \bibinfo{booktitle}{\emph{IEEE International Conference on Computer Vision}}.
\newblock


\bibitem[\protect\citeauthoryear{He, Zhang, Ren, and Sun}{He
  et~al\mbox{.}}{2016}]%
        {he2016deep}
\bibfield{author}{\bibinfo{person}{Kaiming He}, \bibinfo{person}{Xiangyu
  Zhang}, \bibinfo{person}{Shaoqing Ren}, {and} \bibinfo{person}{Jian Sun}.}
  \bibinfo{year}{2016}\natexlab{}.
\newblock \showarticletitle{Deep residual learning for image recognition}. In
  \bibinfo{booktitle}{\emph{IEEE Conference on Computer Vision and Pattern
  Recognition}}.
\newblock


\bibitem[\protect\citeauthoryear{Hendershott, Zhang, Zhao, and
  Zheng}{Hendershott et~al\mbox{.}}{2021}]%
        {hendershott2021fintech}
\bibfield{author}{\bibinfo{person}{Terrence Hendershott},
  \bibinfo{person}{Xiaoquan Zhang}, \bibinfo{person}{J~Leon Zhao}, {and}
  \bibinfo{person}{Zhiqiang Zheng}.} \bibinfo{year}{2021}\natexlab{}.
\newblock \showarticletitle{FinTech as a Game Changer: Overview of Research
  Frontiers}.
\newblock \bibinfo{journal}{\emph{Information Systems Research}}
  \bibinfo{volume}{32}, \bibinfo{number}{1} (\bibinfo{year}{2021}),
  \bibinfo{pages}{1--17}.
\newblock


\bibitem[\protect\citeauthoryear{Hew, Leong, Tan, Ooi, and Lee}{Hew
  et~al\mbox{.}}{2019}]%
        {hew2019age}
\bibfield{author}{\bibinfo{person}{Jun-Jie Hew}, \bibinfo{person}{Lai-Ying
  Leong}, \bibinfo{person}{Garry Wei-Han Tan}, \bibinfo{person}{Keng-Boon Ooi},
  {and} \bibinfo{person}{Voon-Hsien Lee}.} \bibinfo{year}{2019}\natexlab{}.
\newblock \showarticletitle{The age of mobile social commerce: An Artificial
  Neural Network analysis on its resistances}.
\newblock \bibinfo{journal}{\emph{Technological Forecasting and Social Change}}
   \bibinfo{volume}{144} (\bibinfo{year}{2019}), \bibinfo{pages}{311--324}.
\newblock


\bibitem[\protect\citeauthoryear{Hochreiter and Schmidhuber}{Hochreiter and
  Schmidhuber}{1997}]%
        {hochreiter1997long}
\bibfield{author}{\bibinfo{person}{Sepp Hochreiter} {and}
  \bibinfo{person}{J{\"u}rgen Schmidhuber}.} \bibinfo{year}{1997}\natexlab{}.
\newblock \showarticletitle{Long short-term memory}.
\newblock \bibinfo{journal}{\emph{Neural computation}} \bibinfo{volume}{9},
  \bibinfo{number}{8} (\bibinfo{year}{1997}), \bibinfo{pages}{1735--1780}.
\newblock


\bibitem[\protect\citeauthoryear{Jabeur, Sadaaoui, Sghaier, and Aloui}{Jabeur
  et~al\mbox{.}}{2020}]%
        {jabeur2020machine}
\bibfield{author}{\bibinfo{person}{Sami~Ben Jabeur}, \bibinfo{person}{Amir
  Sadaaoui}, \bibinfo{person}{Asma Sghaier}, {and} \bibinfo{person}{Riadh
  Aloui}.} \bibinfo{year}{2020}\natexlab{}.
\newblock \showarticletitle{Machine learning models and cost-sensitive decision
  trees for bond rating prediction}.
\newblock \bibinfo{journal}{\emph{Journal of the Operational Research Society}}
  \bibinfo{volume}{71}, \bibinfo{number}{8} (\bibinfo{year}{2020}),
  \bibinfo{pages}{1161--1179}.
\newblock


\bibitem[\protect\citeauthoryear{Jeong and Kim}{Jeong and Kim}{2019}]%
        {jeong2019improving}
\bibfield{author}{\bibinfo{person}{Gyeeun Jeong} {and}
  \bibinfo{person}{Ha~Young Kim}.} \bibinfo{year}{2019}\natexlab{}.
\newblock \showarticletitle{Improving financial trading decisions using deep
  Q-learning: Predicting the number of shares, action strategies, and transfer
  learning}.
\newblock \bibinfo{journal}{\emph{Expert Systems with Applications}}
  \bibinfo{volume}{117} (\bibinfo{year}{2019}), \bibinfo{pages}{125--138}.
\newblock


\bibitem[\protect\citeauthoryear{Kim and Won}{Kim and Won}{2018}]%
        {kim2018forecasting}
\bibfield{author}{\bibinfo{person}{Ha~Young Kim} {and}
  \bibinfo{person}{Chang~Hyun Won}.} \bibinfo{year}{2018}\natexlab{}.
\newblock \showarticletitle{Forecasting the volatility of stock price index: A
  hybrid model integrating LSTM with multiple GARCH-type models}.
\newblock \bibinfo{journal}{\emph{Expert Systems with Applications}}
  \bibinfo{volume}{103} (\bibinfo{year}{2018}), \bibinfo{pages}{25--37}.
\newblock


\bibitem[\protect\citeauthoryear{Kim}{Kim}{2012}]%
        {kim2012ensemble}
\bibfield{author}{\bibinfo{person}{Myoung-Jong Kim}.}
  \bibinfo{year}{2012}\natexlab{}.
\newblock \showarticletitle{Ensemble learning with support vector machines for
  bond rating}.
\newblock \bibinfo{journal}{\emph{Journal of Intelligence and Information
  Systems}} \bibinfo{volume}{18}, \bibinfo{number}{2} (\bibinfo{year}{2012}),
  \bibinfo{pages}{29--45}.
\newblock


\bibitem[\protect\citeauthoryear{Kingma and Ba}{Kingma and Ba}{2014}]%
        {kingma2014adam}
\bibfield{author}{\bibinfo{person}{Diederik~P Kingma} {and}
  \bibinfo{person}{Jimmy Ba}.} \bibinfo{year}{2014}\natexlab{}.
\newblock \showarticletitle{Adam: A method for stochastic optimization}.
\newblock \bibinfo{journal}{\emph{arXiv preprint arXiv:1412.6980}}
  (\bibinfo{year}{2014}).
\newblock


\bibitem[\protect\citeauthoryear{Kong, Tao, and Fu}{Kong et~al\mbox{.}}{2017}]%
        {kong2017deep}
\bibfield{author}{\bibinfo{person}{Yu Kong}, \bibinfo{person}{Zhiqiang Tao},
  {and} \bibinfo{person}{Yun Fu}.} \bibinfo{year}{2017}\natexlab{}.
\newblock \showarticletitle{Deep sequential context networks for action
  prediction}. In \bibinfo{booktitle}{\emph{IEEE Conference on Computer Vision
  and Pattern Recognition}}.
\newblock


\bibitem[\protect\citeauthoryear{Lai, Chang, Yang, and Liu}{Lai
  et~al\mbox{.}}{2018}]%
        {lai2018modeling}
\bibfield{author}{\bibinfo{person}{Guokun Lai}, \bibinfo{person}{Wei-Cheng
  Chang}, \bibinfo{person}{Yiming Yang}, {and} \bibinfo{person}{Hanxiao Liu}.}
  \bibinfo{year}{2018}\natexlab{}.
\newblock \showarticletitle{Modeling long-and short-term temporal patterns with
  deep neural networks}. In \bibinfo{booktitle}{\emph{The 41st International
  ACM SIGIR Conference on Research \& Development in Information Retrieval}}.
\newblock


\bibitem[\protect\citeauthoryear{Lee}{Lee}{2007}]%
        {lee2007application}
\bibfield{author}{\bibinfo{person}{Young-Chan Lee}.}
  \bibinfo{year}{2007}\natexlab{}.
\newblock \showarticletitle{Application of support vector machines to corporate
  credit rating prediction}.
\newblock \bibinfo{journal}{\emph{Expert Systems with Applications}}
  \bibinfo{volume}{33}, \bibinfo{number}{1} (\bibinfo{year}{2007}),
  \bibinfo{pages}{67--74}.
\newblock


\bibitem[\protect\citeauthoryear{Li, Jin, Xuan, Zhou, Chen, Wang, and Yan}{Li
  et~al\mbox{.}}{2019}]%
        {li2019enhancing}
\bibfield{author}{\bibinfo{person}{Shiyang Li}, \bibinfo{person}{Xiaoyong Jin},
  \bibinfo{person}{Yao Xuan}, \bibinfo{person}{Xiyou Zhou},
  \bibinfo{person}{Wenhu Chen}, \bibinfo{person}{Yu-Xiang Wang}, {and}
  \bibinfo{person}{Xifeng Yan}.} \bibinfo{year}{2019}\natexlab{}.
\newblock \showarticletitle{Enhancing the locality and breaking the memory
  bottleneck of transformer on time series forecasting}.
\newblock \bibinfo{journal}{\emph{Advances in Neural Information Processing
  Systems}} (\bibinfo{year}{2019}).
\newblock


\bibitem[\protect\citeauthoryear{Li{\'e}bana-Cabanillas, Marinkovic, de~Luna,
  and Kalinic}{Li{\'e}bana-Cabanillas et~al\mbox{.}}{2018}]%
        {liebana2018predicting}
\bibfield{author}{\bibinfo{person}{Francisco Li{\'e}bana-Cabanillas},
  \bibinfo{person}{Veljko Marinkovic}, \bibinfo{person}{Iviane~Ramos de Luna},
  {and} \bibinfo{person}{Zoran Kalinic}.} \bibinfo{year}{2018}\natexlab{}.
\newblock \showarticletitle{Predicting the determinants of mobile payment
  acceptance: A hybrid SEM-neural network approach}.
\newblock \bibinfo{journal}{\emph{Technological Forecasting and Social Change}}
   \bibinfo{volume}{129} (\bibinfo{year}{2018}), \bibinfo{pages}{117--130}.
\newblock


\bibitem[\protect\citeauthoryear{Li{\'e}bana-Cabanillas, Marinkovi{\'c}, and
  Kalini{\'c}}{Li{\'e}bana-Cabanillas et~al\mbox{.}}{2017}]%
        {liebana2017sem}
\bibfield{author}{\bibinfo{person}{Francisco Li{\'e}bana-Cabanillas},
  \bibinfo{person}{Veljko Marinkovi{\'c}}, {and} \bibinfo{person}{Zoran
  Kalini{\'c}}.} \bibinfo{year}{2017}\natexlab{}.
\newblock \showarticletitle{A SEM-neural network approach for predicting
  antecedents of m-commerce acceptance}.
\newblock \bibinfo{journal}{\emph{International Journal of Information
  Management}} \bibinfo{volume}{37}, \bibinfo{number}{2}
  (\bibinfo{year}{2017}), \bibinfo{pages}{14--24}.
\newblock


\bibitem[\protect\citeauthoryear{Lim, Zohren, and Roberts}{Lim
  et~al\mbox{.}}{2020}]%
        {lim2020recurrent}
\bibfield{author}{\bibinfo{person}{Bryan Lim}, \bibinfo{person}{Stefan Zohren},
  {and} \bibinfo{person}{Stephen Roberts}.} \bibinfo{year}{2020}\natexlab{}.
\newblock \showarticletitle{Recurrent neural filters: Learning independent
  bayesian filtering steps for time series prediction}. In
  \bibinfo{booktitle}{\emph{International Joint Conference on Neural
  Networks}}.
\newblock


\bibitem[\protect\citeauthoryear{Mourelatos, Alexakos, Amorgianiotis, and
  Likothanassis}{Mourelatos et~al\mbox{.}}{2018}]%
        {mourelatos2018financial}
\bibfield{author}{\bibinfo{person}{Marios Mourelatos},
  \bibinfo{person}{Christos Alexakos}, \bibinfo{person}{Thomas Amorgianiotis},
  {and} \bibinfo{person}{Spiridon Likothanassis}.}
  \bibinfo{year}{2018}\natexlab{}.
\newblock \showarticletitle{Financial indices modelling and trading utilizing
  deep learning techniques: the ATHENS SE FTSE/ASE large cap use case}. In
  \bibinfo{booktitle}{\emph{Innovations in Intelligent Systems and
  Applications}}.
\newblock


\bibitem[\protect\citeauthoryear{Nikolaev, Tino, and Smirnov}{Nikolaev
  et~al\mbox{.}}{2013}]%
        {nikolaev2013time}
\bibfield{author}{\bibinfo{person}{Nikolay Nikolaev}, \bibinfo{person}{Peter
  Tino}, {and} \bibinfo{person}{Evgueni Smirnov}.}
  \bibinfo{year}{2013}\natexlab{}.
\newblock \showarticletitle{Time-dependent series variance learning with
  recurrent mixture density networks}.
\newblock \bibinfo{journal}{\emph{Neurocomputing}}  \bibinfo{volume}{122}
  (\bibinfo{year}{2013}), \bibinfo{pages}{501--512}.
\newblock


\bibitem[\protect\citeauthoryear{Nur-E-Arefin and Mahmud}{Nur-E-Arefin and
  Mahmud}{2020}]%
        {nur2020comparative}
\bibfield{author}{\bibinfo{person}{Md Nur-E-Arefin} {and}
  \bibinfo{person}{Mohammad~Sultan Mahmud}.} \bibinfo{year}{2020}\natexlab{}.
\newblock \showarticletitle{A Comparative Study of Machine Learning Classifiers
  for Credit Card Fraud Detection}.
\newblock \bibinfo{journal}{\emph{International Journal of Innovative
  Technology and Interdisciplinary Sciences}} \bibinfo{volume}{3},
  \bibinfo{number}{1} (\bibinfo{year}{2020}), \bibinfo{pages}{395--406}.
\newblock


\bibitem[\protect\citeauthoryear{Oord, Dieleman, Zen, Simonyan, Vinyals,
  Graves, Kalchbrenner, Senior, and Kavukcuoglu}{Oord et~al\mbox{.}}{2016}]%
        {oord2016wavenet}
\bibfield{author}{\bibinfo{person}{Aaron van~den Oord}, \bibinfo{person}{Sander
  Dieleman}, \bibinfo{person}{Heiga Zen}, \bibinfo{person}{Karen Simonyan},
  \bibinfo{person}{Oriol Vinyals}, \bibinfo{person}{Alex Graves},
  \bibinfo{person}{Nal Kalchbrenner}, \bibinfo{person}{Andrew Senior}, {and}
  \bibinfo{person}{Koray Kavukcuoglu}.} \bibinfo{year}{2016}\natexlab{}.
\newblock \showarticletitle{Wavenet: A generative model for raw audio}.
\newblock \bibinfo{journal}{\emph{arXiv preprint arXiv:1609.03499}}
  (\bibinfo{year}{2016}).
\newblock


\bibitem[\protect\citeauthoryear{Paszke, Gross, Massa, Lerer, Bradbury, Chanan,
  Killeen, Lin, Gimelshein, Antiga, et~al\mbox{.}}{Paszke
  et~al\mbox{.}}{2019}]%
        {paszke2019pytorch}
\bibfield{author}{\bibinfo{person}{Adam Paszke}, \bibinfo{person}{Sam Gross},
  \bibinfo{person}{Francisco Massa}, \bibinfo{person}{Adam Lerer},
  \bibinfo{person}{James Bradbury}, \bibinfo{person}{Gregory Chanan},
  \bibinfo{person}{Trevor Killeen}, \bibinfo{person}{Zeming Lin},
  \bibinfo{person}{Natalia Gimelshein}, \bibinfo{person}{Luca Antiga},
  {et~al\mbox{.}}} \bibinfo{year}{2019}\natexlab{}.
\newblock \showarticletitle{Pytorch: An imperative style, high-performance deep
  learning library}.
\newblock \bibinfo{journal}{\emph{arXiv preprint arXiv:1912.01703}}
  (\bibinfo{year}{2019}).
\newblock


\bibitem[\protect\citeauthoryear{Psaradellis and Sermpinis}{Psaradellis and
  Sermpinis}{2016}]%
        {psaradellis2016modelling}
\bibfield{author}{\bibinfo{person}{Ioannis Psaradellis} {and}
  \bibinfo{person}{Georgios Sermpinis}.} \bibinfo{year}{2016}\natexlab{}.
\newblock \showarticletitle{Modelling and trading the US implied volatility
  indices. Evidence from the VIX, VXN and VXD indices}.
\newblock \bibinfo{journal}{\emph{International Journal of Forecasting}}
  \bibinfo{volume}{32}, \bibinfo{number}{4} (\bibinfo{year}{2016}),
  \bibinfo{pages}{1268--1283}.
\newblock


\bibitem[\protect\citeauthoryear{Puh and Brki{\'c}}{Puh and Brki{\'c}}{2019}]%
        {puh2019detecting}
\bibfield{author}{\bibinfo{person}{Maja Puh} {and} \bibinfo{person}{Ljiljana
  Brki{\'c}}.} \bibinfo{year}{2019}\natexlab{}.
\newblock \showarticletitle{Detecting credit card fraud using selected machine
  learning algorithms}. In \bibinfo{booktitle}{\emph{International Convention
  on Information and Communication Technology, Electronics and
  Microelectronics}}.
\newblock


\bibitem[\protect\citeauthoryear{Raga and Raga}{Raga and Raga}{2019}]%
        {raga2019early}
\bibfield{author}{\bibinfo{person}{Rodolfo~C Raga} {and}
  \bibinfo{person}{Jennifer~D Raga}.} \bibinfo{year}{2019}\natexlab{}.
\newblock \showarticletitle{Early prediction of student performance in blended
  learning courses using deep neural networks}. In
  \bibinfo{booktitle}{\emph{International Symposium on Educational
  Technology}}.
\newblock


\bibitem[\protect\citeauthoryear{Rangapuram, Seeger, Gasthaus, Stella, Wang,
  and Januschowski}{Rangapuram et~al\mbox{.}}{2018}]%
        {rangapuram2018deep}
\bibfield{author}{\bibinfo{person}{Syama~Sundar Rangapuram},
  \bibinfo{person}{Matthias~W Seeger}, \bibinfo{person}{Jan Gasthaus},
  \bibinfo{person}{Lorenzo Stella}, \bibinfo{person}{Yuyang Wang}, {and}
  \bibinfo{person}{Tim Januschowski}.} \bibinfo{year}{2018}\natexlab{}.
\newblock \showarticletitle{Deep state space models for time series
  forecasting}.
\newblock \bibinfo{journal}{\emph{Advances in Neural Information Processing
  Systems}} (\bibinfo{year}{2018}).
\newblock


\bibitem[\protect\citeauthoryear{Salinas, Flunkert, Gasthaus, and
  Januschowski}{Salinas et~al\mbox{.}}{2020}]%
        {salinas2020deepar}
\bibfield{author}{\bibinfo{person}{David Salinas}, \bibinfo{person}{Valentin
  Flunkert}, \bibinfo{person}{Jan Gasthaus}, {and} \bibinfo{person}{Tim
  Januschowski}.} \bibinfo{year}{2020}\natexlab{}.
\newblock \showarticletitle{DeepAR: Probabilistic forecasting with
  autoregressive recurrent networks}.
\newblock \bibinfo{journal}{\emph{International Journal of Forecasting}}
  \bibinfo{volume}{36}, \bibinfo{number}{3} (\bibinfo{year}{2020}),
  \bibinfo{pages}{1181--1191}.
\newblock


\bibitem[\protect\citeauthoryear{Saunders and Allen}{Saunders and
  Allen}{2010}]%
        {saunders2010credit}
\bibfield{author}{\bibinfo{person}{Anthony Saunders} {and}
  \bibinfo{person}{Linda Allen}.} \bibinfo{year}{2010}\natexlab{}.
\newblock \bibinfo{booktitle}{\emph{Credit risk management in and out of the
  financial crisis: new approaches to value at risk and other paradigms}}.
  Vol.~\bibinfo{volume}{528}.
\newblock \bibinfo{publisher}{John Wiley \& Sons}.
\newblock


\bibitem[\protect\citeauthoryear{Sezer, Gudelek, and Ozbayoglu}{Sezer
  et~al\mbox{.}}{2020}]%
        {sezer2020financial}
\bibfield{author}{\bibinfo{person}{Omer~Berat Sezer},
  \bibinfo{person}{Mehmet~Ugur Gudelek}, {and} \bibinfo{person}{Ahmet~Murat
  Ozbayoglu}.} \bibinfo{year}{2020}\natexlab{}.
\newblock \showarticletitle{Financial time series forecasting with deep
  learning: A systematic literature review: 2005--2019}.
\newblock \bibinfo{journal}{\emph{Applied Soft Computing}}
  \bibinfo{volume}{90} (\bibinfo{year}{2020}), \bibinfo{pages}{106181}.
\newblock


\bibitem[\protect\citeauthoryear{Si, Li, Ding, and Rao}{Si
  et~al\mbox{.}}{2017}]%
        {si2017multi}
\bibfield{author}{\bibinfo{person}{Weiyu Si}, \bibinfo{person}{Jinke Li},
  \bibinfo{person}{Peng Ding}, {and} \bibinfo{person}{Ruonan Rao}.}
  \bibinfo{year}{2017}\natexlab{}.
\newblock \showarticletitle{A multi-objective deep reinforcement learning
  approach for stock index future’s intraday trading}. In
  \bibinfo{booktitle}{\emph{International Symposium on Computational
  Intelligence and Design}}.
\newblock


\bibitem[\protect\citeauthoryear{Tan, Steinbach, and Kumar}{Tan
  et~al\mbox{.}}{2016}]%
        {tan2016introduction}
\bibfield{author}{\bibinfo{person}{Pang-Ning Tan}, \bibinfo{person}{Michael
  Steinbach}, {and} \bibinfo{person}{Vipin Kumar}.}
  \bibinfo{year}{2016}\natexlab{}.
\newblock \bibinfo{booktitle}{\emph{Introduction to data mining}}.
\newblock \bibinfo{publisher}{Pearson Education India}.
\newblock


\bibitem[\protect\citeauthoryear{Tran, Balasubramanian, and Hoai}{Tran
  et~al\mbox{.}}{2021}]%
        {tran2021progressive}
\bibfield{author}{\bibinfo{person}{Vinh Tran}, \bibinfo{person}{Niranjan
  Balasubramanian}, {and} \bibinfo{person}{Minh Hoai}.}
  \bibinfo{year}{2021}\natexlab{}.
\newblock \showarticletitle{Progressive Knowledge Distillation For Early Action
  Recognition}. In \bibinfo{booktitle}{\emph{IEEE International Conference on
  Image Processing}}.
\newblock


\bibitem[\protect\citeauthoryear{Vargas, De~Lima, and Evsukoff}{Vargas
  et~al\mbox{.}}{2017}]%
        {vargas2017deep}
\bibfield{author}{\bibinfo{person}{Manuel~R Vargas},
  \bibinfo{person}{Beatriz~SLP De~Lima}, {and} \bibinfo{person}{Alexandre~G
  Evsukoff}.} \bibinfo{year}{2017}\natexlab{}.
\newblock \showarticletitle{Deep learning for stock market prediction from
  financial news articles}. In \bibinfo{booktitle}{\emph{IEEE International
  Conference on Computational Intelligence and Virtual Environments for
  Measurement Systems and Applications}}.
\newblock


\bibitem[\protect\citeauthoryear{Vaswani, Shazeer, Parmar, Uszkoreit, Jones,
  Gomez, Kaiser, and Polosukhin}{Vaswani et~al\mbox{.}}{2017}]%
        {vaswani2017attention}
\bibfield{author}{\bibinfo{person}{Ashish Vaswani}, \bibinfo{person}{Noam
  Shazeer}, \bibinfo{person}{Niki Parmar}, \bibinfo{person}{Jakob Uszkoreit},
  \bibinfo{person}{Llion Jones}, \bibinfo{person}{Aidan~N Gomez},
  \bibinfo{person}{{\L}ukasz Kaiser}, {and} \bibinfo{person}{Illia
  Polosukhin}.} \bibinfo{year}{2017}\natexlab{}.
\newblock \showarticletitle{Attention is all you need}. In
  \bibinfo{booktitle}{\emph{Advances in Neural Information Processing
  Systems}}.
\newblock


\bibitem[\protect\citeauthoryear{Villuendas-Rey, Rey-Bengur{\'\i}a,
  Ferreira-Santiago, Camacho-Nieto, and
  Y{\'a}{\~n}ez-M{\'a}rquez}{Villuendas-Rey et~al\mbox{.}}{2017}]%
        {villuendas2017naive}
\bibfield{author}{\bibinfo{person}{Yenny Villuendas-Rey},
  \bibinfo{person}{Carmen~F Rey-Bengur{\'\i}a}, \bibinfo{person}{{\'A}ngel
  Ferreira-Santiago}, \bibinfo{person}{Oscar Camacho-Nieto}, {and}
  \bibinfo{person}{Cornelio Y{\'a}{\~n}ez-M{\'a}rquez}.}
  \bibinfo{year}{2017}\natexlab{}.
\newblock \showarticletitle{The na{\"\i}ve associative classifier (NAC): a
  novel, simple, transparent, and accurate classification model evaluated on
  financial data}.
\newblock \bibinfo{journal}{\emph{Neurocomputing}}  \bibinfo{volume}{265}
  (\bibinfo{year}{2017}), \bibinfo{pages}{105--115}.
\newblock


\bibitem[\protect\citeauthoryear{Wang, Smola, Maddix, Gasthaus, Foster, and
  Januschowski}{Wang et~al\mbox{.}}{2019}]%
        {wang2019deep}
\bibfield{author}{\bibinfo{person}{Yuyang Wang}, \bibinfo{person}{Alex Smola},
  \bibinfo{person}{Danielle Maddix}, \bibinfo{person}{Jan Gasthaus},
  \bibinfo{person}{Dean Foster}, {and} \bibinfo{person}{Tim Januschowski}.}
  \bibinfo{year}{2019}\natexlab{}.
\newblock \showarticletitle{Deep factors for forecasting}. In
  \bibinfo{booktitle}{\emph{International Conference on Machine Learning}}.
\newblock


\bibitem[\protect\citeauthoryear{Xia, Liu, Da, and Xie}{Xia
  et~al\mbox{.}}{2018}]%
        {xia2018novel}
\bibfield{author}{\bibinfo{person}{Yufei Xia}, \bibinfo{person}{Chuanzhe Liu},
  \bibinfo{person}{Bowen Da}, {and} \bibinfo{person}{Fangming Xie}.}
  \bibinfo{year}{2018}\natexlab{}.
\newblock \showarticletitle{A novel heterogeneous ensemble credit scoring model
  based on bstacking approach}.
\newblock \bibinfo{journal}{\emph{Expert Systems with Applications}}
  \bibinfo{volume}{93} (\bibinfo{year}{2018}), \bibinfo{pages}{182--199}.
\newblock


\bibitem[\protect\citeauthoryear{Xing, Pei, and Philip}{Xing
  et~al\mbox{.}}{2009}]%
        {xing2009early}
\bibfield{author}{\bibinfo{person}{Zhengzheng Xing}, \bibinfo{person}{Jian
  Pei}, {and} \bibinfo{person}{S~Yu Philip}.} \bibinfo{year}{2009}\natexlab{}.
\newblock \showarticletitle{Early prediction on time series: A nearest neighbor
  approach}. In \bibinfo{booktitle}{\emph{International Joint Conference on
  Artificial Intelligence}}.
\newblock


\bibitem[\protect\citeauthoryear{Zhao, Liang, Yu, Wang, and Zhao}{Zhao
  et~al\mbox{.}}{2019}]%
        {zhao2019asynchronous}
\bibfield{author}{\bibinfo{person}{Lei Zhao}, \bibinfo{person}{Huiying Liang},
  \bibinfo{person}{Daming Yu}, \bibinfo{person}{Xinming Wang}, {and}
  \bibinfo{person}{Gansen Zhao}.} \bibinfo{year}{2019}\natexlab{}.
\newblock \showarticletitle{Asynchronous Multivariate time series early
  prediction for ICU transfer}. In \bibinfo{booktitle}{\emph{International
  Conference on Intelligent Medicine and Health}}.
\newblock


\bibitem[\protect\citeauthoryear{Zhou, Zhang, Peng, Zhang, Li, Xiong, and
  Zhang}{Zhou et~al\mbox{.}}{2021}]%
        {zhou2021informer}
\bibfield{author}{\bibinfo{person}{Haoyi Zhou}, \bibinfo{person}{Shanghang
  Zhang}, \bibinfo{person}{Jieqi Peng}, \bibinfo{person}{Shuai Zhang},
  \bibinfo{person}{Jianxin Li}, \bibinfo{person}{Hui Xiong}, {and}
  \bibinfo{person}{Wancai Zhang}.} \bibinfo{year}{2021}\natexlab{}.
\newblock \showarticletitle{Informer: Beyond efficient transformer for long
  sequence time-series forecasting}. In \bibinfo{booktitle}{\emph{AAAI
  Conference on Artificial Intelligence}}.
\newblock


\bibitem[\protect\citeauthoryear{Zhou, Pan, Hu, Tang, and Zhao}{Zhou
  et~al\mbox{.}}{2018}]%
        {zhou2018stock}
\bibfield{author}{\bibinfo{person}{Xingyu Zhou}, \bibinfo{person}{Zhisong Pan},
  \bibinfo{person}{Guyu Hu}, \bibinfo{person}{Siqi Tang}, {and}
  \bibinfo{person}{Cheng Zhao}.} \bibinfo{year}{2018}\natexlab{}.
\newblock \showarticletitle{Stock market prediction on high-frequency data
  using generative adversarial nets}.
\newblock \bibinfo{journal}{\emph{Mathematical Problems in Engineering}}
  \bibinfo{volume}{2018} (\bibinfo{year}{2018}).
\newblock


\bibitem[\protect\citeauthoryear{Zhou, Han, Xu, Jiang, and Zhang}{Zhou
  et~al\mbox{.}}{2019}]%
        {zhou2019long}
\bibfield{author}{\bibinfo{person}{Yu-Long Zhou}, \bibinfo{person}{Ren-Jie
  Han}, \bibinfo{person}{Qian Xu}, \bibinfo{person}{Qi-Jie Jiang}, {and}
  \bibinfo{person}{Wei-Ke Zhang}.} \bibinfo{year}{2019}\natexlab{}.
\newblock \showarticletitle{Long short-term memory networks for CSI300
  volatility prediction with Baidu search volume}.
\newblock \bibinfo{journal}{\emph{Concurrency and Computation: Practice and
  Experience}} \bibinfo{volume}{31}, \bibinfo{number}{10}
  (\bibinfo{year}{2019}), \bibinfo{pages}{e4721}.
\newblock


\end{thebibliography}

\end{document}